\documentclass{article}
\usepackage{arxiv}
\usepackage[utf8]{inputenc} 
\usepackage[T1]{fontenc}    
\usepackage{url}            
\usepackage{booktabs}       
\usepackage{nicefrac}         
\usepackage{graphicx}
\usepackage[numbers]{natbib}
\usepackage{doi}
\usepackage{graphicx}
\usepackage{tikz}
\usepackage{soul}
\usepackage{comment}
\mathchardef\mhyphen="2D
\usepackage{amssymb, amsmath}
\usepackage{color}
\usepackage{multirow}
\usepackage{graphicx,wrapfig,lipsum}
\usepackage{subfigure}
\usepackage{booktabs} 
\usepackage{dsfont}
\usepackage{wrapfig}
\usepackage{mwe}
\usetikzlibrary{backgrounds}
\usepackage{caption}
\usetikzlibrary{intersections}
\usetikzlibrary{shapes.geometric}
\usetikzlibrary{spy}
\usepackage{color, colortbl}
\usepackage{xcolor}
\usepackage[accsupp]{axessibility} 
\definecolor{Gray}{gray}{0.9}
\usepackage[accsupp]{axessibility}  
\usepackage[noend]{algpseudocode}
\usepackage{algorithm}
\usepackage{algorithmicx}
\usepackage{etoolbox}

\title{DASS: Differentiable Architecture Search For Sparse Neural Networks}

\date{} 				

\author{
   Hamid Mousavi\\
	Mälardalen University\\
	\texttt{seyedhamidreza.mousavi@mdu.se} \\
	\And
	\hspace{1mm}Mohammad Loni\\
	Mälardalen University\\
	\texttt{mohammad.loni@mdu.se} \\
	\AND
	\hspace{1mm}Mina Alibeigi\\
	\texttt{mina.alibeigi@zenseact.com}
	\AND
	\hspace{1mm}Masoud Daneshtalab \\
        Mälardalen University\\
	\texttt{masoud.daneshtalab@mdu.se}
}

\date{}

\renewcommand{\shorttitle}{DASS: Differentiable Architecture Search For Sparse Neural Networks}
\begin{document}
\maketitle
\begin{abstract}
The deployment of Deep Neural Networks (DNNs) on edge devices is hindered by the substantial gap between performance requirements and available processing power.
While recent research has made significant strides in developing pruning methods to build a sparse network for reducing the computing overhead of DNNs, there remains considerable accuracy loss, especially at high pruning ratios.
We find that the architectures designed for dense networks by differentiable architecture search methods are ineffective when pruning mechanisms are applied to them.
The main reason is that the current method does not support sparse architectures in their search space and uses a search objective that is made for dense networks and does not pay any attention to sparsity. 

In this paper, we propose a new method to search for sparsity-friendly neural architectures. We do this by adding two new sparse operations to the search space and modifying the search objective.
We propose two novel parametric \texttt{SparseConv} and \texttt{SparseLinear} operations in order to expand the search space to include sparse operations.
In particular, these operations make a flexible search space due to using sparse parametric versions of linear and convolution operations.
The proposed search objective lets us train the architecture based on the sparsity of the search space operations.
Quantitative analyses demonstrate that our search architectures outperform those used in the state-of-the-art sparse networks on the CIFAR-10 and ImageNet datasets.
In terms of performance and hardware effectiveness, DASS increases the accuracy of the sparse version of MobileNet-v2 from 73.44\% to 81.35\% (+7.91\% improvement) with 3.87$\times$ faster inference time.
\end{abstract}
\section{Introduction}
\label{sec:introduction}

Deep Neural Networks (DNNs) provide an excellent avenue for obtaining the maximum feature extraction capacities required to resolve highly complex computer vision tasks \citep{ren2015faster,he2016deep,voulodimos2018deep,qin2018convolutional}.
There is an increasing demand for DNNs to become more efficient in order to be deployed on extremely resource-constrained edge devices.
However, DNNs are not intrinsically tailored for the limited computing and memory capacities of tiny edge devices, prohibiting their deployment in such applications  \citep{gholami2020ai,loni2021faststereonet,lin2022device,lin2020mcunet,loni2020deepmaker}.

To democratize DNN acceleration, a variety of optimization approaches have been proposed, including network pruning \citep{sehwag2020hydra,liang2021pruning,zhang2021training,diao2023pruning}, efficient architecture design \citep{loni2021faststereonet,lin2020mcunet}, network quantization \citep{loni2021TAS,bulat2020bats,kim2020learning}, knowledge distillation \citep{hinton2015distilling,gou2021knowledge}, and low-rank decomposition \citep{jaderberg2014speeding}.
Particularly, network pruning is known to provide remarkable computational and memory savings by removing redundant weight parameters in the unstructured scenario \citep{sehwag2020hydra,azarian2020learned,liang2021pruning,zhang2021training,halabi2022data}, and the entire filter in the structured scenario \citep{he2018soft,he2019filter,he2020learning,zhuang2020neuron,zhang2022advancing}.
Recently, unstructured pruning methods reported to provide extreme network size reductions.
The state-of-the-art unstructured pruning methods\citep{sehwag2020hydra} provide up to 99\% pruning ratio which is an excellent scenario for tiny edge devices. 

Nevertheless, these methods suffer from a substantial accuracy drop, hampering them from being applied in practice ($\approx$19\% accuracy drop for MobileNet-v2 compared to dense one \citep{sehwag2020hydra}).
Current pruning methods use handcrafted architectures designed without concern about sparsity. \citep{azarian2020learned,liang2021pruning,zhang2021training,zhuang2020neuron,sehwag2020hydra}.
We hypothesize that the backbone architecture may not be optimal for scenarios with extreme pruning ratios.
Instead, we can learn more efficient backbone architectures adaptable to pruning techniques by exploring the space of sparse networks.

Neural Architecture Search (NAS) has achieved great success in the automated designing of high-performance DNN architectures.
Differentiable architecture search (DARTS) methods \citep{liu2018darts,ye2022b,ye2022beta} is a popular NAS method that uses a gradient-based search algorithm to expedite the search speed. 
Motivated by the promising results of NAS, we came up with the idea of designing customized backbone architectures compatible with pruning methods.
Nevertheless, the search space of current DARTS algorithms comprises dense convolution and linear operations that are incapable of exploring the correct backbone for pruning.
To demonstrate this issue, we first prune 99\% of the weights from the best architecture designed by NAS method \citep{liu2018darts} with base search space without regard for sparsity.
Disappointingly, after applying the pruning method to the final architecture, it performs poorly with up to $\approx$21\% accuracy loss in compression by DASS that extends the search space by sparse operations. (Section~\ref{sec:motivation}).
This failure is due to a lack of support for specific sparse network characteristics leading to low generalization performance.
Based on the above hypothesis and empirical observations, we formulate a search space that includes sparse and dense operations.
Therefore, the original convolution and linear operations in the search space of the NAS have been extended by parametric \texttt{SparseConv} and \texttt{SparseLinear} operations, respectively. 
Moreover, to make a consistency between the proposed search space and search objective function, we modify the bi-level optimization problem to take sparsity into account.
In this way, the search process tries to find the best sparse operation by optimizing both architecture and pruning parameters.    
This modification creates a complex bi-level optimization problem.
To tackle this difficulty, we split the complex bi-level optimization into two simple bi-level optimization problems and solve them.

\begin{figure}[t]
\centering
\captionsetup{justification=centering}
    \begin{center}
		    \includegraphics[width = \columnwidth]{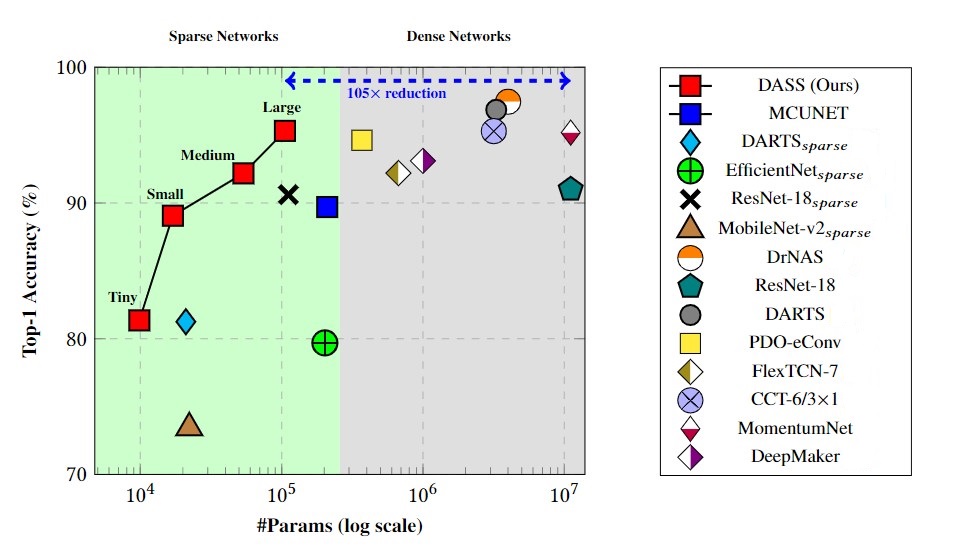}
	        \end{center}	
\caption{Top-1 accuracy (\%) vs. number of network parameters (\#Params) trained on CIFAR-10 for various sparse and dense architectures.} 
\label{fig:introduction}
\end{figure}

We show explicitly integrating pruning into the search procedure can lead to finding sparse network architectures with significant accuracy improvement. 
In Fig.~\ref{fig:introduction}, we compare the CIFAR-10 Top-1 accuracy and the number of parameters of the found architecture by DASS with the state-of-the-art sparse (unstructured pruning) and dense networks.
Results show the designed architecture by DASS outperforms all competing architectures that employ the pruning method.
DASS-Small demonstrates its consistent effectiveness by achieving 15\%, 10\%, and 8\% accuracy improvement over MobileNet-v2$_{sparse}$ \citep{sandler2018mobilenetv2}, EfficientNet-v2$_{sparse}$ \citep{tan2019efficientnet}, and DARTS$_{sparse}$ \citep{liu2018darts}, respectively.
In addition, compared to networks with similar accuracy, DASS-Large has a significant reduction in network complexity (\#Params) by 3.5$\times$, 30.0$\times$, 105.2$\times$ over PDO-eConv \citep{shen2020pdo}, CCT-6/3$\times$1 \citep{romero2021flexconv}, and MomentumNet \citep{sander2021momentum}, respectively.
Section~\ref{sec:experiments} provides a comprehensive experimental study to evaluate different aspects of DASS. 
Our main contributions are summarized as follows:

\begin{enumerate}
    \item We perform extensive experiments to identify the limitations of applying pruning with extreme pruning ratios to the dense architecture as a post-processing step. 
    \item We define a new search space by extending the base search space with a new set of parametric operations (\texttt{SparseConv} and \texttt{SparseLinear}) to consider the sparse operations in the search space.
    \item We modify the bi-level optimization problem to be consistent with the new search space and propose a three-step gradient-based algorithm to split the complex bi-level problem and learn architecture parameters, network weights, and pruning parameters.
\end{enumerate}

\section{Related Work}
\label{sec:related_work}

\subsection{Neural Architecture Search and DARTS Variants}
\label{sec:related_work:NAS}
Neural Architecture Search (NAS) has recently attracted remarkable attention by relieving human experts from the laborious effort of designing neural networks.  
Early NAS methods mainly utilized evolutionary-based \citep{real2019regularized,loni2019neuropower,loni2020deepmaker,liu2021survey} or reinforcement-learning-based methods \citep{zoph2018learning,zoph2016neural,jaafra2019reinforcement}.
Despite the efficiencies of handcrafted designs, they require tremendous computing resources. For example, the proposed method in \citep{zoph2018learning} evaluates 20,000 neural candidates across 500 NVIDIA\textsuperscript{®} P100 GPUs over four days.
One-shot architecture search methods \citep{brock2017smash,guo2020single,bender2018understanding} have been proposed to identify optimal neural architectures within a few GPU days ($>$1 GPU day \citep{ren2021comprehensive}).
In particular, Differentiable Architecture Search (DARTS)  \citep{liu2018darts,ye2022b,ye2022beta} is a variation of one-shot NAS methods that relaxes the search space to be continuous and differentiable. The detailed description of DARTS can be found in Section~\ref{sec:preliminaries:DARTS}.
Despite the broad successes of DARTS in advancing NAS applicability, achieving optimal results remains a challenge for real-world problems.
Many subsequent works investigate some of these challenges by focusing on (i) increasing search speed \citep{hundt2019sharpdarts,siddiqui2021operation}, (ii) improving generalization performance \citep{chen2020drnas,wang2021rethinking}, (iii) addressing the robustness issues \citep{zela2019understanding,yue2020effective,hosseini2021dsrna}, (iv) reducing quantization error \citep{kim2020learning,loni2021TAS}, and (v) designing hardware-aware architectures \citep{jin2019rc,lee2021help,cai2019once}.
On the other hand, few works attempt to prune the search space by removing inferior network operations \citep{laube2019prune,noy2020asap,hong2021dropnas,ding2022nap}.
These works utilized the pruning mechanism to progressively remove some operations from the search space.  
Unlike them, our method aims to extend the search space to improve the performance of the sparse network by searching for the best operations with sparse weight structures. 
Technically, our method extends the search space by adding the parametric sparse version of convolution and linear operations to find the best sparse architecture.   
Therefore, there is a lack of research on sparse weight parameters when designing neural architectures.
DASS searches for the operations that are most effective for sparse weight parameters in order to achieve higher generalizing performance.

\subsection{Network Pruning}
\label{sec:related_work:pruning}

Network pruning is an effective method for reducing the size of DNNs, enabling them to be effectively deployed on devices with limited resource capacity.
Prior works on network pruning can be classified into two categories: structured and unstructured pruning methods. 
The purpose of structured pruning is to remove redundant channels or filters to preserve the entire structure of weight tensors with dimension reduction \citep{he2018soft,he2019filter,li2019compressing,he2020learning,zhuang2020neuron,guan2022dais}.
While structured pruning is famous for hardware acceleration, it sacrifices a certain degree of flexibility as well as weight sparsity \citep{liu2018rethinking}. 

On the other hand, unstructured pruning methods offer superior flexibility and compression rate by removing parameters with the least impact on the network accuracy from the weight tensors \citep{han2015learning,li2016pruning,he2018soft,liu2018rethinking,frankle2018lottery,sehwag2020hydra,azarian2020learned,liang2021pruning,zhang2021training}. 
In general, unstructured pruning entails three stages to make a sparse network, including (i) pre-training, (ii) pruning, and (iii) fine-tuning. 
Prior unstructured pruning methods used various criteria to select the lowest pruning weight parameters.
\citep{lecun1990optimal,hassibi1993second} pruned weight parameters based on the second-derivative values of the loss function.
Several studies proposed to remove the weight parameters below a fixed pruning threshold, regardless of the training objective \citep{han2015learning,li2016pruning,frankle2018lottery,zhou2019deconstructing,ye2019adversarial,gui2019model}.
To address the limitation of fixed thresholding methods, \citep{azarian2020learned,Kusupati20} proposed layer-wise trainable thresholds to determine the optimal value for each layer separately.
The lottery-ticket hypothesis \citep{frankle2018lottery,burkholz2021existence,chen2022data} is a different line of the method that identifies the pruning mask for an initialized CNN and trains the resulting sparse model from scratch without changing the pruning mask.
 HYDRA \citep{sehwag2020hydra} formulate the pruning objective as empirical risk minimization and integrate it with the training objective. 
Unlike other methods, optimization-based pruning criteria improve the performance of sparse networks in comparison to other metrics.
Despite the success of optimization-based pruning in achieving a significant compression rate, classification accuracy is compromised, notably when the pruning ratio is extremely high (up to 99\%).
We show that the main reason for this issue is due to the non-optimal backbone architecture.
We extend the search space of DASS by parametric sparse operations and formulate pruning as an empirical risk minimization problem and integrate it into the bi-level optimization problem to find the best sparse network.

\section{Preliminaries}
\label{sec:preliminaries}

\subsection{Differentiable Architecture Search}
\label{sec:preliminaries:DARTS}

Differentiable Architecture Search (DARTS) \citep{liu2018darts} is a NAS method that significantly reduces the search cost by relaxing the search space to be continuous and differentiable.
DARTS cell template is represented by a Directed Acyclic Graph (DAG) containing $N$ intra-nodes.
The edge $(i,j)$ between two nodes is associated with an operation $o^{(i,j)}$ (e.g., skip connection or $3\times3$ max-pooling) within $\mathcal{O}$ search space.
Eq.~\ref{eq3} computes the output of intermediate nodes. 
\begin{equation}\label{eq3}
    \bar{o}^{(i,j)} (x^{(i)}) = \sum_{o\in \mathcal{O}}\frac{exp\big(\alpha_o^{(i,j)}\big)}{\sum_{o^\prime \in \mathcal{O}}exp \big(\alpha_{o^\prime}^{(i,j)}\big)}\cdot o (x^{(i)})
\end{equation}

where $\mathcal{O}$ and $\alpha_{o}^{(i,j)}$ denote the set of all candidate operations and the selection probability of $o$, respectively.
The output node in the cell is the concatenation of all intermediate nodes.
DARTS optimizes architecture parameters ($\alpha$) and network weights ($\theta$) with the following bi-level objective function:

\begin{equation}\label{eq4}
    \mathop{\mathrm{min}}_{\alpha} \mathcal{L}_{val}(\theta^{\star},\alpha) \;\; s.t. \; \theta^{\star} = \mathop{\mathrm{argmin}}_{\theta} \mathcal{L}_{train}(\theta,\alpha)
\end{equation}
where 
\begin{equation*}
 \mathcal{L}_{train} = \frac{\sum_{(\boldsymbol{x},y) \in (X_{train},Y_{train})} l(\theta,\boldsymbol{x},y)}{|X_{train}|}    
\end{equation*}
and
\begin{equation*}
    \mathcal{L}_{val} = \frac{\sum_{(\boldsymbol{x},y) \in (X_{val},Y_{val})} l(\theta,\boldsymbol{x},y)}{|X_{val}|}
\end{equation*}
The operation with the largest $\alpha_o$ is selected for each edge.
$X_{train}$ and $Y_{train}$ represent the training dataset and corresponding labels, respectively.
Similarly, the validation dataset and labels are indicated by $X_{val}$ and $Y_{val}$, respectively.
After the search process has been completed, the final architecture is re-trained from scratch to obtain maximum accuracy.

\subsection{Unstructured Pruning} 
\label{sec:preliminaries:HYDRA}

Pruning is considered unstructured if it removes low-importance parameters from the weight tensors and makes sparse ones. \citep{liu2018rethinking}.
This paper uses the unstructured network pruning method based on optimization criteria to provide higher flexibility and an extreme compression rate compared to structured pruning methods.
The pruning method includes three main optimization stages: (i) pre-training: training the network on the target dataset, (ii) pruning: pruning unimportant weights from the pre-trained network, and (iii) fine-tuning: the sparse network is re-trained to recover its original accuracy.
For the pruning stage, we consider an optimization-based method  with the following steps: 
First, we define the pruning parameters that show the importance of each weight of the network ($s^0$) and initialize them according to Eq.~\ref{eq2}.

\begin{equation}\label{eq2}
    s_i^0 \propto \frac{1}{\max (|\theta_{pre,i}|)} \times \theta_{pre,i}
\end{equation}

where $\theta_{pre,i}$ denotes the weight of $i_{th}$ layer in the pre-trained network. 
Next, to learn the pruning parameters $(\hat{s})$, we formulate the optimization problem as Eq.~\ref{eq1}, which is then solved by the stochastic gradient descent (SGD) \citep{Goodfellow-et-al-2016}.

\begin{equation} \label{eq1}
    \hat{s} = \mathop{\mathrm{argmin}}_{s}  \mathbb{E}_{(x, y)\sim D} \big[\mathcal{L}_{prune}(\theta_{pre},s, x, y)\big]\\
\end{equation}

$\theta_{pre}$ and $\mathbb{E}$ refer to the pre-trained network parameters and mathematical expectation, respectively.
By solving this optimization problem, we are able to determine the effect of each weight parameter on the loss function and, consequently, the accuracy of the network.  
Finally, we convert the floating values of the pruning parameters to a binary mask based on selecting top-$k$ weights with the highest magnitude of pruning parameters.

\begin{table}[b]
\centering
\captionsetup{justification=centering}
\resizebox{\columnwidth}{!}{
\begin{tabular}{cc}
		\includegraphics[width = \columnwidth]{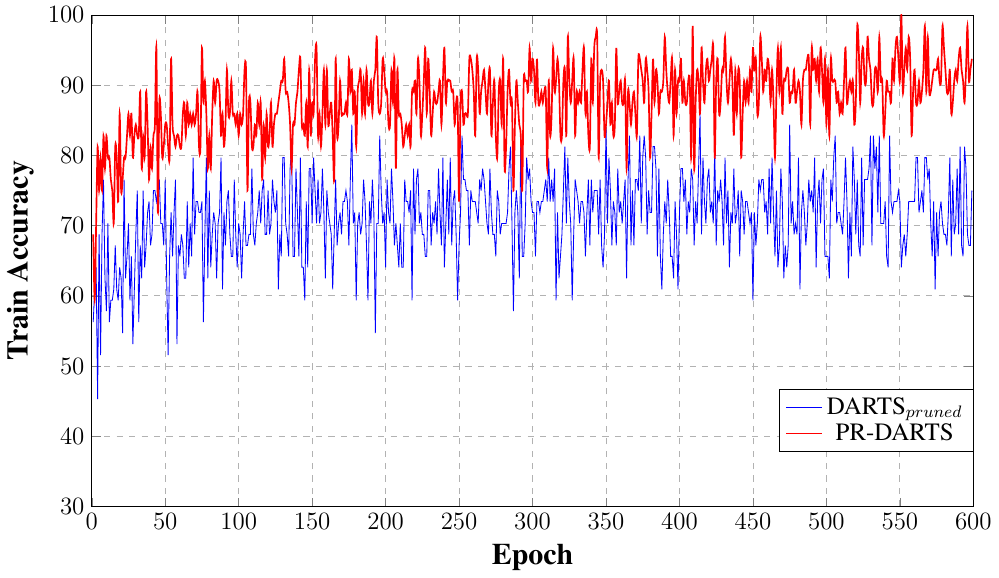}
&
		\includegraphics[width = \columnwidth]{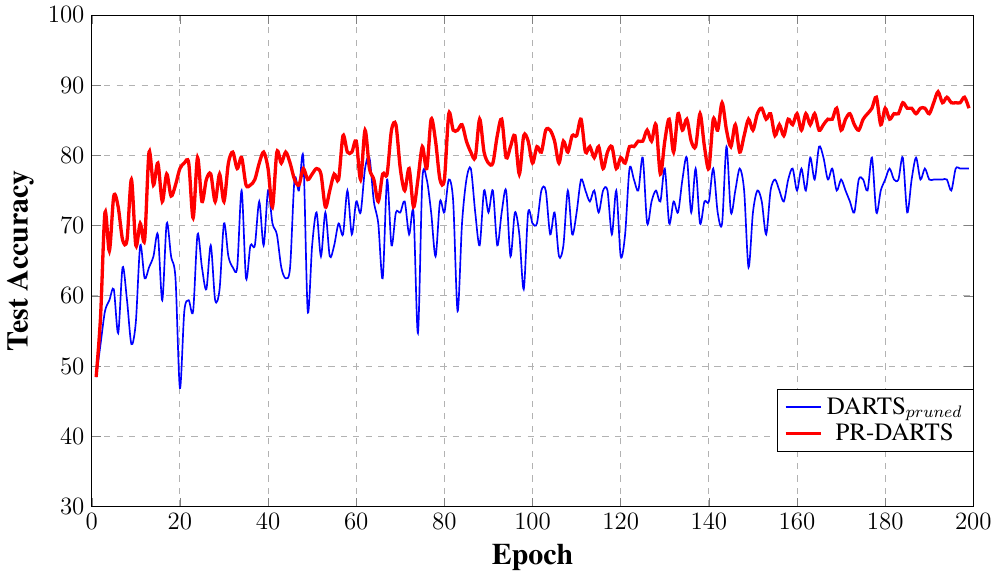}
\\
\textbf{\Large  (a) Train} &  \textbf{\Large (b) Test}
\end{tabular}
}
\captionof{figure}{Comparison of DASS-Small and DARTS$_{sparse}$ on CIFAR-10 for (a) train and (b) test learning curves.} 
\label{fig:motivation}
\end{table}

\section{Research Motivation}
\label{sec:motivation}

\begin{figure}[b]
\captionsetup{justification=centering}
	\begin{center}
		\includegraphics[width = 0.5\columnwidth]{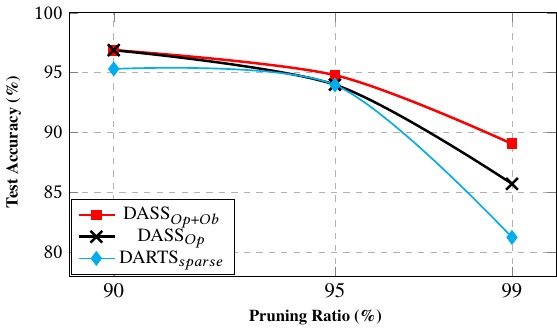}
	\end{center}	
\caption{DASS$_{Op+Ob}$ vs. DARTS$_{sparse}$ and DASS with only adding sparse operations to the search space (DASS$_{Op}$).} 
\label{fig:motivation:motivatoin_new}
\end{figure}

The dense network architectures that were originally designed using conventional NAS methods are inaccurate when integrated with pruning methods, particularly at high pruning ratios.
To demonstrate this assertion, we first apply the unstructured pruning method explained in section~\ref{sec:preliminaries:HYDRA} to the best architecture designed by DARTS \citep{liu2018darts} for CIFAR-10 and generate a sparse network.
We call this solution DARTS$_{sparse}$.
Then, we compare the performance of the sparse architecture designed by DASS with DARTS$_{sparse}$.
Fig.~\ref{fig:motivation} illustrates the train and test accuracy curves for DASS and DARTS$_{sparse}$ architectures trained on the CIFAR-10 dataset.
Disappointingly, the network designed by DARTS$_{sparse}$ results in reduced test accuracy.
This implies that the dense backbone architectures designed by NAS methods without considering sparsity are ineffective (DASS delivers 8\% higher test accuracy compared to DARTS$_{sparse}$).
According to our investigations, we find two issues involved in the training failure of DARTS$_{sparse}$: (i) DARTS does not support sparse operations in its search space, and (ii) DARTS optimizes the search objective without considering sparsity into account.
Section~\ref{sec:method:PrunedConv} addresses the first issue, while the second issue is addressed in Section~\ref{sec:method:PR-DARTSObjective}.
We investigate DASS in two modes to demonstrate the significance of including sparse operations and reformulating the objective function based on sparsity. The first mode extends the search space with sparse operations solely (DASS$_{Op}$) and does not optimize the pruning parameters, while the second mode adds sparsity to the optimization process and optimizes the architecture and pruning parameters in a bi-level optimization problem. (DASS$_{Op+Ob}$).
Fig.~\ref{fig:motivation:motivatoin_new} indicates the test accuracy for DASS$_{Op}$, DARTS$_{sparse}$ and (DASS$_{Op+Ob}$) architectures with various pruning ratios.
As results show, DASS$_{Op}$ has $\approx$ 3.4\% lower accuracy compared to DASS$_{Op+Ob}$ and $\approx$ 4.47\% higher accuracy compared to DARTS$_{sparse}$.
In conclusion, extending the search space with proposed sparse operations (our first contribution) in DASS produces a better architecture than DARTS$_{sparse}$, but combining it with the sparsity-based optimization objective (our second contribution) enhances the performance.

\section{DASS method}
\label{sec:method}

\subsection{DASS: Overview}
\label{sec:method:overview}

We propose DASS, a differentiable architecture search method for sparse neural networks.
DASS at first extends the search space of the NAS with parametric sparse operations.
Then it modifies the bi-level optimization problem to learn the architecture, weights, and pruning parameters.
DASS employs a three-step approach to solve the complicated bi-level optimization problem, which consists of (1) \textit{Pre-training}: Find the best dense architecture (pruning parameters equal to zero) from the search space and pre-train it (2) \textit{Pruning and sparse Architecture Design}: Find the best pruning mask (optimizing pruning parameters) and update the architecture parameters based on the sparse weights and finally (3) \textit{Fine-tuning}: we re-train the sparse architecture to achieve the maximum classification performance.

\subsection{DASS Search Space}
\label{sec:method:PrunedConv}

\begin{figure}[h]
	\begin{center}
		\includegraphics[width = \columnwidth]{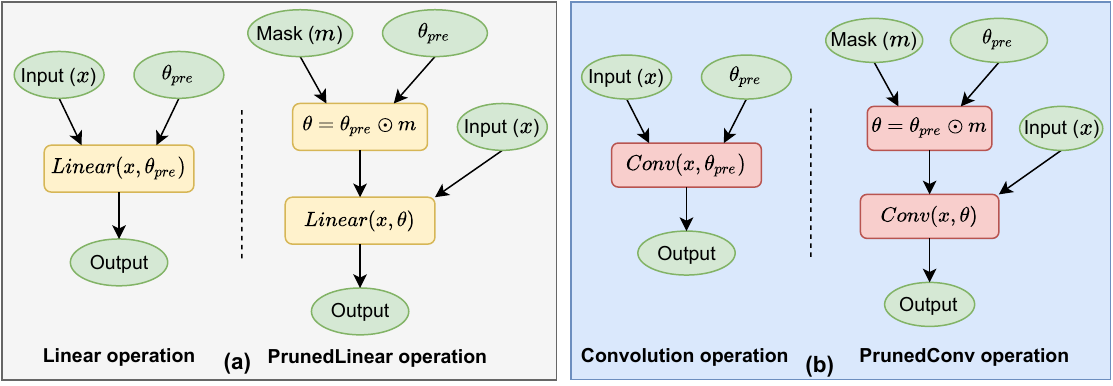}
	\end{center}	
\caption{Illustrating the (a) \texttt{SparseLinear} and (b) \texttt{SparseConv} operations.}
\label{fig:PrunedConv}
\end{figure}

\begin{figure}[b]
\captionsetup{justification=centering}
	\begin{center}
		\includegraphics[width = \columnwidth]{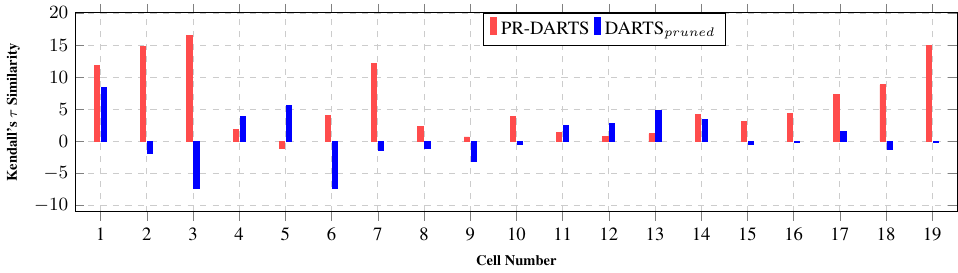}
	\end{center}	
\caption{Comparing the Kendall’s $\tau$ similarity metric of architectures designed by both DARTS$_{sparse}$ and DASS methods with high-performance dense architecture .} 
\label{fig:method:kendalTau}
\end{figure}

To support sparse operations, DASS proposes the parametric sparse version of convolution and linear operations called \texttt{SparseConv} and \texttt{SparseLinear}, respectively.
These operations have a sparsity mask ($m$) to remove redundant weight parameters from the network. 
Fig.~\ref{fig:PrunedConv} illustrates the functionality of these two operations.
In addition, table~\ref{sec:method:tab:operations} summarizes the operations of the DASS search space.
\begin{table}[hbpt]
\caption{Operations of the DASS search space.}
\label{sec:method:tab:operations}
\resizebox{\columnwidth}{!}{%
		\begin{tabular}{cccccc}\hline
		\textbf{Operation}	   & Separable & Dilated & Max & Average & Skip \\
		\textbf{Type} & Sparse Convolution & sparse Convolution& Pooling & pooling & connect\\
		\hline
		\textbf{Kernel Size} & $3\times3$, $5\times5$ & $3\times3$, $5\times5$& $3\times3$& $3\times3$ & N/A \\
		\hline
		\end{tabular}}
\end{table} 
To empirically investigate the efficiency of the proposed sparse search space, we compare the similarity of the feature maps of high-performance dense architecture (with a large number of parameters) with the sparse architecture discovered by DASS and the architecture designed from the original search space  DARTS$_{spasre}$  methods. 
We use Kendall’s $\tau$ \citep{abdi2007kendall} metric to measure the similarity between output feature maps. The $\tau$ correlation coefficient returns a value between -1 and 1. To present the outcome more clearly, we scale up these values between -100 and 100. 
Closer values to 100 indicate stronger positive similarity between the feature maps.
Fig.~\ref{fig:method:kendalTau} summarizes the results. 
Our observations reveal a  similarity between DASS feature maps and dense architecture  (up to 16\%). On the other hand, the correlation between DARTS$_{sparse}$ and dense architecture  is insignificant.
Therefore, it shows that the architecture designed by DASS based on new search space can extract features more similar to high-performance dense architecture while DARTS$_{sparse}$ that use dense search space lost important features after pruning.
The level of similarity is not very high because DASS is a sparse network with a pruning ratio of 99\%. However, it can demonstrate that DASS retrieves useful features.

\subsection{DASS Search objective}
\label{sec:method:PR-DARTSObjective}

DASS aims to search for the optimal architecture parameters ($\alpha^\star$) to minimize the validation loss of the sparse network weight parameters.
Thus, to consistent search objective with proposed sparse search space. we formulate the entire search objective as a complex bi-level optimization problem:

\begin{equation}\label{eq5}
    \begin{split}
        \alpha^\star=&\mathop{\mathrm{min}}_{\alpha}(\mathcal{L}_{val}(\hat{\theta}(\alpha),\alpha))\\
        s.t. \;\;\;\; &
         \begin{cases} 
           \theta^{\star}(\alpha) = \mathop{\mathrm{argmin}}_{\theta} \mathcal{L}_{train}(\theta,\alpha)\\
            \hat{m} =\mathop{\mathrm{argmin}}_{m\in\{0,1\}^N}  \big[\mathcal{L}_{prune}(\theta^\star(\alpha) \odot m,\alpha)\big] \\
           \hat{\theta}(\alpha) = \theta^\star(\alpha)\odot \hat{m}.
        \end{cases}
    \end{split}
\end{equation}

Where $m$ denotes the binary pruning mask parameters. 
This formulation learns the architecture parameters based on the sparse weight parameters. 
However, Eq.~\ref{eq5} is not a straightforward bi-level optimization problem because the lower-level problem consists of two optimization problems. To overcome this challenge,  We break the search objective down into three distinct steps. Thus, the problem is transformed into two bi-level optimization problems to determine the optimal architecture parameters for dense and sparse weights and an optimization problem to fine-tune the weight parameters.
In addition, the lower-level optimization problem consists of a discrete optimization problem for pruning masks. 

Section~\ref{sec:method:OptimizationAlgorithm} proposes a multi-step optimization algorithm to solve the optimization problem and handle the discrete optimization problem by converting it to a continuous optimization problem.
\subsection{Optimization Algorithm}
\label{sec:method:OptimizationAlgorithm}
\subsubsection*{Step 1: pre-train (learn $\theta^\star_{pre}$ and $\alpha^*_{pre}$)}
\label{sec:method:OptimizationAlgorithm:pre-train}

In this step, we break the  Eq.~\ref{eq5} into a bi-level optimization problem to find the best dense architecture. 
This pre-training is necessary for the next step which learn pruning mask parameters and modifying the sparse architecture.
\begin{equation}
    \begin{split}
        \alpha^\star_{pre}=&\mathop{\mathrm{min}}_{\alpha_{pre}}(\mathcal{L}_{val}(\theta^*_{pre}(\alpha_{pre}),\alpha_{pre}))\\
        s.t. \;\;\;\; & 
        \theta^{\star}_{pre}(\alpha_{pre}) = \mathop{\mathrm{argmin}}_{\theta_{pre}} \mathcal{L}_{train}(\theta_{pre},\alpha_{pre})
    \end{split}
\end{equation}

The first-order approximation technique use to update $\theta^\star_{pre}$ and $\alpha_{pre}$ alternately using gradient descent \citep{liu2018darts}.
%
\subsubsection*{Step 2: prune (learn $\hat{m}$ and $\alpha^*_{prune}$)}
\label{sec:method:OptimizationAlgorithm:pruning}

To make the search process aware of the sparsity mechanism, we need to solve another bi-level optimization problem that alternately updates the pruning mask and architecture parameters. 
Pruning mask parameters are binary values. Therefore, learning the mask parameters ($m$) is a challenging binary optimization problem. 
We solve this binary optimization problem by introducing floating point pruning parameters $s$ and initializing them. Then we use SGD to solve the optimization problem and find the best floating-point pruning mask parameters.
%
%
%
%
Finally, Based on the values of pruning parameters, we select the top-$k$ weight parameters with the highest values and assign one value to them.
%
%
This step aims to jointly learn architecture parameters $\alpha_{prune}$ and mask parameters $\hat{m}$ to consider sparsity in learning architecture parameters.
Therefore, we use another bi-level optimization problem:
\begin{equation}
    \begin{split}
        \alpha^\star_{prune}=&\mathop{\mathrm{min}}_{\alpha_{prune}}(\mathcal{L}_{val}(\theta^*_{pre}\odot \hat{m}(\alpha_{prune}),\alpha_{prune}))\\
        s.t. \;\;\;\; & 
        \hat{s}(\alpha_{prune}) = \mathop{\mathrm{argmin}}_{s} \mathcal{L}_{prune}(\theta^*_{pre},\alpha_{prune},s), \\ &
        \hat{m}(\alpha_{prune}) = \mathds{1}(|\hat{s}(\alpha_{prune})|>|\hat{s}(\alpha_{prune})|_k)
    \end{split}
\end{equation}
similar to step 1, the first-order approximation method is used to alternately update $\hat{m}$ and $\alpha_{prune}$ by
gradient descent.
\subsubsection*{Step 3: fine-tune (learn $\hat{\theta}$)}
\label{sec:method:OptimizationAlgorithm:fine-tune}
In the fine-tuning step, we update the non-zero weight parameters using SGD for the best sparse architecture to improve the network accuracy (Eq.~\ref{eq8}).
\begin{equation}\label{eq8}
    \hat{\theta}_{t+1} = \hat{\theta}_{t} - \eta_{\hat{\theta}}\nabla_{\hat{\theta}} \mathcal{L}_{fine\mhyphen tune}(\hat{\theta}_{t} \odot \hat{m},\alpha^\star_{prune})
\end{equation}
where $\eta_{\hat{\theta}}$ and $\mathcal{L}_{fine\mhyphen tune}$ denote the learning rate and the loss function for the fine-tuning step.
%
%
%

\begin{figure}[t]
	\begin{center}
		\includegraphics[width = \columnwidth]{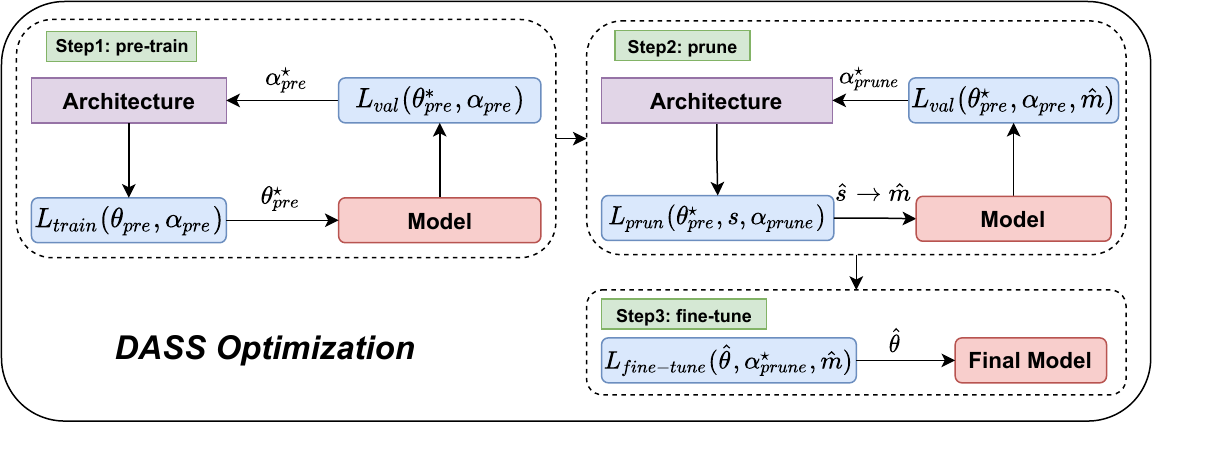}
	\end{center}	
\caption{The overview of the proposed optimization algorithm to find architecture parameters based on the sparse weight parameters. It consists of three main steps: 1) pre-training: search dense architecture 2) pruning: search sparse architecture 3) fine-tuning: re-train best sparse architecture.}
\label{fig:method_overview}
\end{figure}

\begin{figure}[b]
    \centering
    \begin{minipage}{0.47\textwidth}
        	\begin{center}
		    \includegraphics[width = \columnwidth]{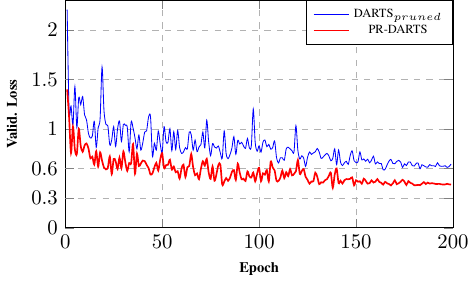}
	        \end{center}	
        \caption{Comparing learning curves (validation loss) of DASS and DARTS$_{sparse}$ on the searched architectures trained with the CIFAR-10 dataset.}
        \label{fig:motivation:PrunedConv}
    \end{minipage}\hfill
    \begin{minipage}{0.47\textwidth}
        	\begin{center}
		    \includegraphics[width = \columnwidth]{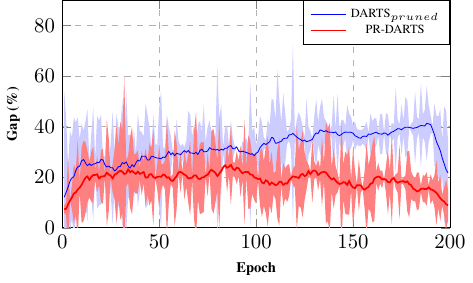}
	        \end{center}	
        \caption{Comparing the generalization gap of DASS and DARTS$_{sparse}$ over the CIFAR-10 dataset. The lower values for the generalization gap are better. }
        \label{fig:motivation:generalizationGap}
    \end{minipage}
\end{figure}

We show that the proposed three-step optimization algorithm can solve the complex bilevel problem in Eq.~\ref{eq5} and  finds optimal architecture parameters with higher generalization performance for sparse networks. 
Fig.~\ref{fig:motivation:PrunedConv} compares the learning curves of DASS with DARTS$_{sparse}$ on the CIFAR-10 dataset.
As shown, the DASS optimization algorithm significantly reduces the validation loss for the sparse network.    
Fig.~\ref{fig:motivation:generalizationGap} compares the behavior of the generalization gap (train minus test accuracy) for DASS and DARTS$_{sparse}$.
DASS has a lower generalization gap (up to 22\%), indicating DASS better regularizes the validation loss across all epochs compared to DARTS$_{sparse}$. Algorithm \ref{Supplementary_Material:alg:PR-DARTS} outlines our DASS for the differentiable neural architecture search for sparse neural networks.
\begin{algorithm}[hbpt]
\caption{Search Process of the DASS}
\label{Supplementary_Material:alg:PR-DARTS}
\begin{algorithmic}[1]
\Require{Dataset $D$, loss objectives: $\mathcal{L}_{train}$, $\mathcal{L}_{prune}$, and $L_{fine-tune}$, training iteration $T$}
\Ensure{fine-tuned spasre model}
\Statex \underline{\textit{Step1: Pre-train}}
    \For{$i \gets 1$ to $T$}
    \State keep $\alpha_{pre}^t$ fixed, and obtain $\theta^{t+1}_{pre}$ by gradient descent with $\nabla_{\theta_{pre}} \mathcal{L}_{train}(\theta_{pre}^{t},\alpha^t_{pre})$
    \State keep  $\theta^{t+1}_{pre}$  fixed, and obtain $\alpha_{pre}^{t+1}$ by gradient descent with $\nabla_{\alpha_{pre}} \mathcal{L}_{val}(\theta_{pre}^{t+1},\alpha^t_{pre})$ 
    \EndFor
    \Statex  \underline{\textit{Step2: Prune}}
    \For{$i \gets 1$ to $T$}
    \State keep $\alpha_{prune}^t$ fixed, and obtain $s^{t+1}$ by gradient descent with $\nabla_{s} \mathcal{L}_{prune}(s^{t},\alpha^t_{prune})$
    \State Compute $m^{t+1}=(|s^{t+1}|>|s^{t+1}|_k)$
    \State keep  $m^{t+1}$  fixed, and obtain $\alpha_{prune}^{t+1}$ by gradient descent with $\nabla_{\alpha_{prune}} \mathcal{L}_{val}(\theta^*_{pre}\odot m^{t+1},\alpha^t_{prune})$ 
    \EndFor
    \Statex  \underline{\textit{Step3: fine-tune}}
    \For{$i \gets 1$ to $T$}
    \State keep $\alpha^*_{prune}$ and $\hat{m}$ fixed and obtain $\hat{\theta}^{t+1}$ by  gradient descent with  $\nabla_{\hat{\theta}} \mathcal{L}_{fine-tune}(\hat{\theta^t} \odot \hat{m},\alpha^*_{prune})$
    \EndFor\\
    \Return{Fine-tuned sparse model}
\end{algorithmic}
\end{algorithm}

\section{Experiments}
\label{sec:experiments}

\definecolor{color1}{rgb}{0.392,0.392,1.0}
\definecolor{color1}{rgb}{0.392,0.392,1.0}
\definecolor{color2}{rgb}{0.0,0.0,0.0}
\definecolor{color3}{rgb}{0.203,0.121,0.592}
\definecolor{color4}{rgb}{0.341,0.396,0.454}
\definecolor{color5}{rgb}{0.501,0.313,0.501}
\definecolor{color6}{rgb}{0.313,0.674,0.517}
\definecolor{color7}{rgb}{1.0,0.623,0.952}
\definecolor{color8}{rgb}{0.039,0.741,0.890}
\definecolor{color9}{rgb}{1.0,0.419,0.419}
\definecolor{color10}{rgb}{0.996,0.792,0.341}

\subsection{Experimental Setup}
\label{sec:experiments:experimental_setup}

\textit{1) DATASET}:
To evaluate DASS, we use CIFAR-10 \citep{krizhevsky2009cifar} and ImageNet \citep{krizhevsky2012imagenet} public classification datasets.
For the search process, we split the CIFAR-10 dataset into 30k data points for training and 30k for validation.
We transfer the best-learned cells on CIFAR-10 to ImageNet \citep{liu2018darts} and re-train the final sparse network from scratch.

\textit{2) Details on Searching Networks}:
We create a network with 16 initial channels and eight cells.
Each cell consists of seven nodes equipped with a depth-wise concatenation operation as the output node.
The \texttt{SparseConv} operations follow the ReLU+\texttt{SpasreConv}+Batch Normalization order.
We train the network using SGD for 50 epochs with a batch size of 64 in the DASS pre-train step.
Then, we update the value of pruning and architecture parameters for 20 epochs in the DASS pruning step.
Finally, we fine-tune the network for 200 epochs.
The initial learning rate for the DASS in pre-train, pruning, and fine-tuning steps is 0.025, 0.1, and 0.01, respectively. 
In our experiments, we use the cosine annealing learning rate \citep{loshchilov2016sgdr}. 
We use weight decay=3$\times$10\textsuperscript{-4} and momentum=0.9 in all steps.
The search process takes $\approx$3 GPU-days on a single NVIDIA\textsuperscript{®} RTX A4000 that produces 4.35 Kg $CO_2$.
We compare the sparse architecture design by our method, DASS, with other dense and sparse networks. 
NAS-Bench-101 \citep{ying2019bench} and NAS-Bench-201 \citep{dong2020bench} are examples of NAS algorithm evaluation benchmarks. 
They consist of numerous dense designs and their respective performance.
Due to the fact that they do not support sparse architectures, we cannot evaluate DASS using these benchmarks.
Creating sparse benchmarks for evaluating NAS algorithms is a suggestion for future work.  

\textit{3) DASS variants and Hardware Configuration }:
Table~\ref{tab:ArchitectureConfig} provides the configuration details of the DASS variants.
Each variation is built by stacking a different number of DASS cells and the output channels of the first layer to generate networks for various resource budgets.
Table~\ref{sec:expriments:tab:HW_Spec} presents specifications of hardware devices utilized for evaluating the performance of DASS at inference time.

\begin{table}[hbpt]
  \centering
  \caption{Configuration of the DASS variants. \#Cells: the number of stacked cells. \#Channels: the number of output channels for the first \texttt{SparseConv} operation.} 
  \label{tab:ArchitectureConfig}
    \resizebox{\columnwidth}{!}{\begin{tabular}{ccccc|ccc}
    \hline
    \multirow{2}{*}{\textbf{DASS}}  & \multicolumn{4}{c}{\textbf{CIFAR-10}}& \multicolumn{3}{c}{\textbf{ImageNet}}\\
     & Tiny & Small & Medium & Large & Small & Medium & Large\\ \hline
    \#Cells & 16 &20&12&14&14&15&16 \\ \hline
    \#Channels & 30 &36&86&108&48&86&128 \\ \hline
\end{tabular}}
\end{table}
\begin{table}[hbpt]
  \centering
  \caption{Hardware Specification.}
  \label{sec:expriments:tab:HW_Spec}
  \resizebox{\columnwidth}{!}{
    \begin{tabular}{ccc}
    \hline
    \textbf{Platform} & \textbf{Specification} & \textbf{Value}\\ \hline
 \multirow{6}{*}{\textbf{\rotatebox{40}{\footnotesize	 Search \& Train }}}  &  GPU  & NVIDIA\textsuperscript{®} RTX A4000 (735 MHz)\\
  &  GPU Memory	& 16 GB GDDR6  \\
  &  GPU Compiler	&  cuDNN version 11.1  \\ 
  &  System Memory & 64 GB \\
  &  Operating System & Ubuntu 18.04 \\ 
  &  $CO_2$ Emission/Day $^\dagger$ & 1.45 Kg \\
  \hline 

\multirow{8}{*}{\textbf{\rotatebox{40}{\footnotesize Real Hardware }}}  
&   \multirow{4}{*}{Embedded GPU} & NVIDIA\textsuperscript{®} Jetson TX2 (735 MHz)\\
&  & 256 CUDA Cores\\ \cline{3-3}
&  &NVIDIA\textsuperscript{®} Quadro M1200 (735 MHz)\\
&  & 640 CUDA Cores\\ 
\cline{2-3}
& \multirow{4}{*}{Embedded CPU} & ARM Cortex\textsuperscript{TM}-A7 (1.2 GHz)\\
&  & 4/4 (Cores/Total Thread)\\\cline{3-3}
&  & Intel\textsuperscript{®}i5-3210M Mobile CPU\\
&  & 5/4 (Cores/Total Thread)\\
\cline{1-3}
\multirow{4}{*}{\textbf{\rotatebox{40}{\footnotesize Estimation$\ddagger$ }}}   &   \multirow{2}{*}{Xiaomi Mi9 GPU} & Adreno 640 GPU (750 MHz)\\
&  & 986 GFLOPs FP32 (Single Precision) \\\cline{2-3}
&  \multirow{2}{*}{Myriad VPU} & Intel Movidius NCS2 (700 MHz)\\
&  & 28-nm Co-processor  \\\hline
\multicolumn{3}{c}{$\dagger$ Calculated using the ML $CO_2$ impact framework: \url{https://mlco2.github.io/impact/} \citep{lacoste2019quantifying}}	\\

\multicolumn{3}{c}{$\ddagger$ Performance Estimation using the nn-Meter framework \citep{zhang2021nn}.}
\\ \hline
\end{tabular}}
\end{table}

\subsection{DASS Compared to dense Networks}
\label{sec:experiments:Full-Precision}

Table~\ref{tab:Full-Precision_results} compares the performance of DASS against the state-of-the-art and the state-of-the-practice DNNs.
We select the architecture with the highest accuracy, DrNAS \citep{chen2020drnas}, as the baseline for comparing compression rates.
In comparison with DrNAS \citep{chen2020drnas}, DASS-Large provides 37.73$\times$ and 29.23$\times$ higher network compression rates while delivering a comparable accuracy (less than 2.5\% accuracy loss) on the CIFAR-10 and ImageNet datasets, respectively.
Compared to the best handcrafted designed network \citep{romero2021flexconv} on the CIFAR-10 (CCT-6/3x1), DASS-Large significantly decreases the parameters of the network by 29.9$\times$ with providing slightly higher accuracy.
\begin{table}[htbp]
\centering
\caption{Comparing the DASS method with the state-of-the-art dense networks on the CIFAR-10  and ImageNet datasets.}
\label{tab:Full-Precision_results}
\resizebox{\linewidth}{!}{%
\begin{tabular}{ccc|ccc|cccc}
\hline
\multirow{3}{*}{\textbf{Architecture}} & \multirow{3}{*}{\textbf{Year}}&&\multicolumn{3}{c}{\textbf{CIFAR-10}}& \multicolumn{4}{c}{\textbf{ImageNet}}\\\cline{4-10}
& & \textbf{Search}& \textbf{Top-1} &\textbf{\#Params}& \textbf{ \#Params} & \textbf{Top-1} & \textbf{Top-5} & \textbf{\#Params}& \textbf{ \#Params} \\
& & \textbf{Method}& \textbf{Acc.(\%)} & \textbf{($\times 10^6$)} & \textbf{Compression}&\textbf{Acc.(\%)} &\textbf{Acc.(\%)}& \textbf{($\times 10^6$)} & \textbf{Compression}\\ \hline
ResNet-18$^\ddagger$ \citep{he2016deep} & 2016 &-&  91.0  & 11.1 & -2.77$\times$ & 72.33 & 91.80 & 11.7 & -2.05$\times$  \\ 
PDO-eConv \citep{shen2020pdo} & 2020 &-& 94.62   &  0.37 & +10.81$\times$ & -& -& -& -\\
FlexTCN-7 \citep{romero2021flexconv} & 2021 &-& 92.2  & 0.67 & +5.97$\times$ & -& -&- & -  \\
CCT-6/3x1 \citep{romero2021flexconv} & 2021 &-& 95.29  &  3.17 & +1.26$\times$ & - & -&- &-\\  
MomentumNet \citep{sander2021momentum} & 2021 &-& 95.18  &  11.1 & -2.77$\times$  &- &- & -& -\\ 
DARTS (1$^{st}$ order) \citep{liu2018darts} &2018&gradient& 96.86  & 3.3 & +1.21$\times$ & -&- &- &- \\
DARTS (2$^{nd}$ order) \citep{liu2018darts} &2018 &gradient& 97.24  & 3.3 & +1.21$\times$ & 74.3 & 91.3 &  4.7 & +1.21$\times$\\
SGAS (Cri 1. avg) \citep{li2020sgas} &2020 &gradient& 97.34  & 3.7 & +1.08$\times$ & 75.9 & 92.7 & 5.4& +1.05$\times$\\
SDARTS-RS \citep{chen2020stabilizing} & 2020 &gradient& 97.39  & 3.4 &  +1.17$\times$ & 75.8 & 92.8 & 3.4 & +1.67$\times$\\
DrNAS \citep{chen2020drnas} &2020 & gradient&\textbf{\textcolor{color1}{97.46}}  & 4.0 & 1.0$\times$& \textbf{\textcolor{color1}{76.3}}& 92.9 & 5.7 & 1.0$\times$ \\
 \hline
DASS-Small  & 2022 & gradient& 89.06 &  \textbf{\textcolor{color1}{ 0.017}}  & \textbf{\textcolor{color1}{+235.29$\times$}} & 46.48 & 68.36 &  \textbf{\textcolor{color1}{0.029}} &\textbf{\textcolor{color1}{+196.55$\times$}} \\ 
DASS-Medium  & 2022&gradient & 92.18       &     0.054   & +74.07$\times$ & 68.34 & 82.24 & 0.082 & +69.51$\times$ \\
DASS-Large  & 2022&gradient & 95.31  & 0.106  & +37.73$\times$ & 73.83& 85.94 &0.195 & +29.23$\times$  \\
\hline
\multicolumn{10}{c}{$\dagger$ The baseline for comparing the \#params compressing rate is  DrNAS \citep{chen2020drnas} as the most accurate architecture.}	\\
\multicolumn{10}{c}{$^\ddagger$ ResNet-18 results are trained in {https://github.com/facebook/fb.resnet.torch}{Torch} (July 10, 2018).} \\
\hline
\end{tabular}}
\end{table} 
\subsection{DASS Compared to sparse Networks}
\label{sec:experiments:pruned}
As we focus on improving the accuracy of sparse networks at extremely high pruning ratios, we compare DASS with other sparse networks with the unstructured pruning method at 99\% pruning ratio (Table~\ref{tab:pruned_results}).
In comparison with DARTS$_{sparse}$, DASS-Small yields 7.81\% and 7.81\% higher top-1 accuracies with 1.23$\times$ and 1.05$\times$ reduction in network size on the CIFAR-10 and ImageNet datasets, respectively.
It indicates that the network design based on new search space and sparse objective function finds better sparse architecture. 
In comparison with ResNet-18$_{sparse}$ on the CIFAR-10 dataset, we provide 1.56\% and 4.7\% higher accuracy with 2.08$\times$ and 1.05$\times$ network size reduction for DASS-Medium and DASS-Large, respectively.
Compared to ResNet-18$_{sparse}$ on the ImageNet dataset, DASS-Medium provides 0.76\% higher accuracy with 1.42$\times$ network size reduction.
MCUNET \citep{lin2020mcunet} is a lightweight neural network for microcontrollers. It is designed by a tiny neural architecture search mechanism. 
Compared to MCUNET on the  ImageNet dataset, DASS-Large provides 1\% higher accuracy with 2.89$\times$ network size reduction.
This result shows that only optimizing the size of the filters without considering the sparsity can not generate the best architecture. 
DASS directly search for the best operations in sparse version to design high-performance lightweight network. 
We can conclude that DASS increases sparse networks' accuracy at high pruning ratios compared to NAS-based and handcrafted networks.
\begin{table*}[hbpt]
\caption{Comparing the DASS method with sparse networks on the CIFAR-10 and ImageNet datasets.}
\label{tab:pruned_results}
\resizebox{\linewidth}{!}{%
\begin{tabular}{ccccc|ccccc}
\hline
 \multirow{3}{*}{\textbf{ Architecture}} & \multicolumn{4}{c}{\textbf{CIFAR-10}} & \multicolumn{4}{c}{\textbf{ImageNet}} \\
\cline{2-10}

  & {\textbf{Top-1}} & \textbf{\#Params}   &\textbf{Compression} & \multirow{2}{*}{\textbf{NID$^\ddagger$}} &  {\textbf{Top-1} } & {\textbf{Top-5} } & \textbf{\#Params} & \textbf{Compression} & \multirow{2}{*}{\textbf{NID$^\ddagger$}}
\\ 
 & {\textbf{Acc. (\%)}} & \textbf{($\times10^3$)} & \textbf{Rate$^\dagger$} & & {\textbf{Acc. (\%)}} &
{\textbf{Acc. (\%)}} & \textbf{($\times10^3$)}  & \textbf{Rate$^\dagger$} &
\\ \hline
DARTS$_{sparse}$ \citep{liu2018darts} & 81.25 & 21.0  & 100.47$\times$ & 3.86 & 38.67 & 61.33 & 33.0 & 100$\times$ & 1.11\\ 
MobileNet-v2$_{sparse}$  \citep{sandler2018mobilenetv2} & 73.44 & 22.2  & 95.04$\times$ & 3.30& 17.97 & 36.72 &34.87 & 94.63$\times$  & 0.515\\
 ResNet-18$_{sparse}$  \citep{he2016deep} & 90.62  & 111.6  & 18.90$\times$ & 0.81 & 67.58  &80.86 & 116.84 & 28.24$\times$ & 0.578 \\ 
EfficientNet$_{sparse}$ \citep{tan2019efficientnet} & 79.69 & 202.3  & 10.43$\times$ &0.39 & - & - & - & -& -\\ 
MCUNET \citep{lin2020mcunet} & 89.7  & 210.1 & 15.70 & 0.42 & 72.34 & 84.86 & 562.64 & 5.86$\times$& 0.128\\ \hline
 DASS-Small  &  89.06 &\textbf{\textcolor{color1}{17.0}}  & \textbf{\textcolor{color1}{124.11$\times$}}  & \textbf{\textcolor{color1}{5.23}} & 46.48 & 68.36 &\textbf{\textcolor{color1}{28.94}}  & \textbf{\textcolor{color1}{114.02$\times$}} &\textbf{\textcolor{color1}{1.606}}\\ 
 DASS-Medium &  92.18 &  53.65 & 39.32$\times$ & 1.71 & 68.34 & 82.24& 81.95& 40.26$\times$  &0.841\\
 DASS-Large &  \textbf{\textcolor{color1}{95.31}} &  105.5  & 20$\times$ & 0.90& \textbf{\textcolor{color1}{73.83}}& \textbf{\textcolor{color1}{85.94}}  & 194.6 &16.95$\times$&0.38\\
\hline
\multicolumn{10}{c}{$\dagger$ The baseline for comparing the compressing rate is full-precision and dense DARTS architecture.}	\\
\multicolumn{10}{c}{$\ddagger$ NID = Accuracy/\#Parameters \citep{bianco2018benchmark}. NID measures how efficiently each network uses its parameters.} \\
 \hline
\end{tabular}}
\end{table*}

\subsection{Evaluation of DASS with Various Pruning Ratios}
\label{sec:expriments:pruning_ratio}

Table~\ref{tab:Sensitivity_pruning_ratio} compares DASS and the DARTS$_{sparse}$ method with three different pruning ratios including 90\%, 95\%, and 99\% on the CIFAR-10 dataset. 
DASS achieves 1.57\%, 1.04\%, and 7.8\% higher accuracies with 7\%, 6.9\%, and 23\% network size reduction compared to the DARTS$_{sparse}$ at 90\%, 95\%, and 99\% pruning ratios, respectively. 
Thus, DASS is significantly more effective at extremely higher pruning ratios (99\%) than lower pruning ratios (90\%).

\begin{table}[hbpt]
\centering
\caption{Evaluating the effectiveness of DASS at various pruning ratios.}
\label{tab:Sensitivity_pruning_ratio}
\resizebox{\linewidth}{!}{%
\begin{tabular}{ccccccc}
\hline
\multirow{3}{*}{\textbf{ Architecture}} & 
\multicolumn{2}{c}{\textbf{90\%}} &
\multicolumn{2}{c}{\textbf{95\%}} &
\multicolumn{2}{c}{\textbf{99\%}} 
\\\cline{2-7}
& {\textbf{ Accuracy}} & \textbf{\#Params}& {\textbf{ Accuracy}} & \textbf{\#Params} & {\textbf{ Accuracy}} & \textbf{\#Params }\\
& & $(\times 10^3)$&&$(\times 10^3)$& &$(\times 10^3)$\\
\hline
DARTS$_{sparse}$ & 95.31\% & 421 & 93.75\%  & 210.5& 81.25\% & 21.0\\
DASS-Small & \textbf{ \textcolor{color1}{96.88\% }} & \textbf{\textcolor{color1}{391}} &\textbf{\textcolor{color1}{94.79\% }} &\textbf{\textcolor{color1}{196.75}}&\textbf{\textcolor{color1}{89.06\%}} & \textbf{\textcolor{color1}{17.0}}  \\
\hline
\end{tabular}}
\end{table}

\subsection{DASS Compared to Other Pruning Methods}
\label{sec:experiments:pruning_methods}

Table~\ref{tab:PR-DARTS_other_pruning_methods} compares DASS with state-of-the-art pruning algorithms.
The results indicate that DASS outperforms other pruning algorithms with different backbone architectures on CIFAR-10 and ImageNet datasets.
On CIFAR-10, DASS-Large shows a 1.6\% higher accuracy and 3.8$\times$ reduction in the network size compared to the most accurate results provided by TAS$_{Pruning}$ \citep{dong2019network}.  
DASS-Large also provides 4.68\% accuracy improvement with 38.14$\times$ reduction in the network size over TAS$_{Pruning}$ \citep{dong2019network} on ImageNet.
In light of DASS' higher efficiency compared to other pruning methods, we can conclude that the pruning method was not the only reason for the DASS's effectiveness and it is independent of the pruning algorithm. 

\begin{table*}[hbpt]
  \centering
  \caption{Comparing DASS with other pruning algorithms.}
  \label{tab:PR-DARTS_other_pruning_methods}
  \resizebox{\linewidth}{!}{%
    \begin{tabular}{c|ccc|cccc}
    \hline
    \multirow{3}{*}{\textbf{Pruning Method}}& \multicolumn{3}{c}{\textbf{CIFAR-10}}&\multicolumn{4}{c}{\textbf{ImageNet}}\\ \cline{2-8}
    & \textbf{Backbone}& \textbf{Top-1} & \textbf{\#Params} & \textbf{Backbone}& \textbf{Top-1} & \textbf{Top-5}& \textbf{\#Params}\\ 
    & \textbf{Arch.}& \textbf{Acc.(\%)} & \textbf{($\times 10^6$)} &\textbf{Arch.}& \textbf{Acc.(\%)} & \textbf{Acc.(\%)}& \textbf{($\times 10^6$)}\\ \hline 
    SFP \citep{he2018soft} & \multirow{3}{*}{{\textbf{\rotatebox{40}{\scriptsize	 ResNet-20 }}}}& 92.08 & 0.269 & \multirow{3}{*}{{\textbf{\rotatebox{40}{\scriptsize	 ResNet-18 }}}} &67.10 & 87.78 & 6.46 \\
    FPGM \citep{he2019filter}& & 92.31 & 0.269 & & 68.41 & 88.48 & 6.46 \\
    TAS$_{Pruning}$ \citep{dong2019network} & & 93.16 & 0.232 & & 69.15 & 88.48 & 
    7.40 \\ \hline
    DASS-Small & - & 89.06 & \textbf{\textcolor{color1}{0.017}} & - & 46.48 & 68.36& \textbf{\textcolor{color1}{0.029}} \\
    DASS-Medium & - & 92.18 & 0.054 & - & 68.34 & 82.24& 0.082 \\
    DASS-Large & - & \textbf{\textcolor{color1}{95.31}} & 0.106 & - &  \textbf{\textcolor{color1}{73.83}} & 85.94 & 0.194 \\ \hline
\end{tabular}}
\end{table*}
\subsection{DASS Compared to Quantized Networks}
\label{sec:experiments:quantization}

Network quantization emerged as a promising research direction to reduce the computation of neural networks.
Recently, \citep{kim2020learning,bulat2020bats,loni2021TAS} proposed to integrate the quantization mechanism into the differentiable NAS procedure to improve the performance of quantized networks.
Table~\ref{tab:quantized_results} compares DASS with the best results of NAS-based quantized networks. 
The compression rate is calculated as $\frac{\sum_{l=1}^{L} \#W_l \times 32}{\sum_{l=1}^{L} \#W_l^t \times q}$ where $\#W_l$ and $\#W_l^t$ are the number of weights in layer $l$ for full-precision (32-bit) and quantized network with $q\mhyphen$bit resolution \citep{loni2021TAS}.
DASS-Medium yields 0.24\% and 3.24\% higher accuracies and significantly higher compression rate by 2.7$\times$ and 4.24$\times$ compared to TAS \citep{loni2021TAS} as the most accurate quantized network on the CIFAR-10 and ImageNet datasets, respectively. 

\begin{table*}[hbpt]
\centering
\caption{Comparing the DASS method with quantized networks on CIFAR-10. }
\label{tab:quantized_results}
\resizebox{\linewidth}{!}{%
\begin{tabular}{cc|ccc|cccc}
\hline
\multirow{3}{*}{\textbf{Architecture}} & & \multicolumn{3}{c}{\textbf{CIFAR-10}}& \multicolumn{4}{c}{\textbf{ImageNet}}\\ \cline{3-9}
 & \textbf{\#bits}&\textbf{Top-1}&\textbf{\#Params} & \textbf{Compression} & \textbf{Top-1}&\textbf{Top-5}&\textbf{\#Params} & \textbf{Compression}  \\
 & \textbf{(W/A)$^\ddagger$} & \textbf{Acc.(\%)}&\textbf{($\times 10^6$)}   & Rate$^{\dagger}$ &\textbf{Acc.(\%)}&\textbf{Acc.(\%)}&\textbf{($\times 10^6$)}   & Rate$^{\dagger}$ \\ \hline
 Binary NAS (A) \citep{kim2020learning} & 1/1 & 90.66 & 2.4  &  44.0$\times$ &57.69 &79.89& 5.57&32.74$\times$\\
 TAS  \citep{loni2021TAS} &  2/2 & 91.94 &2.4    & 22.0$\times$ & 65.1&86.3 &5.57 & 16.37$\times$ \\\hline
DASS-Small &  32/32  &  89.06 & \textbf{\textcolor{color1}{0.017}} & \textbf{\textcolor{color1}{194.11$\times$}} & 46.48 & 68.36 & \textbf{\textcolor{color1}{0.029}}& \textbf{\textcolor{color1}{196.55$\times$}}\\ 
DASS-Medium  &  32/32     &  92.18  &  0.054       & 61.11$\times$ & 68.34 & 82.24 & 0.082 & 69.51$\times$\\
DASS-Large  &  32/32      & \textbf{\textcolor{color1}{95.31}} &  0.106  & 31.13$\times$ & \textbf{\textcolor{color1}{73.83}} & 85.94 &0.194 &29.38$\times$\\
\hline     
\multicolumn{9}{c}{$\dagger$ The baseline for comparison is full-precision DARTS with 3.3M and 5.7M parameters for CIFAR-10 and ImageNet.}	\\
\multicolumn{9}{c}{$\ddagger$ (Weights/Activation Function).} \\ \hline
\end{tabular}}
\end{table*} 

\subsection{Hardware Performance Results of DASS}
\label{sec:experiments:hardware}

We extensively study the effectiveness of DASS in the context of hardware efficiency by computing the inference time (latency) of various state-of-the-art sparse networks for a wide range of resource-constrained edge devices on the CIFAR-10 dataset (Fig.~\ref{fig:experiments:hardware_results}). 
The batch size is equal to 1 for all experiments. 
It is worth noting that we did not utilize any simplification techniques, such as \citep{bragagnolo2022simplify}, to compact the sparse filters by fusing weight parameters. 
Our results reveal that the Pareto-frontier of DASS consistently outperforms all other counterparts by a significant margin, especially on CPUs that have very limited parallelism. 
DASS-Tiny as the fastest network improves the accuracy from MobileNet-v2’s 73.44\% to 81.35\% (+7.91\% improvement) and accelerates the inference by up to 3.87$\times$. More importantly, DASS-Tiny runs much faster than DARTS$_{sparse}$ by 1.67-4.74$\times$ with slightly better accuracy. 
Compared to ResNet-18$_{sparse}$ as the closest network to DASS in terms of accuracy, DASS-Medium provides 1.46\% accuracy improvement and up to 1.94$\times$ acceleration on hardware.

\begin{table*}[t]
\captionsetup{justification=centering}
\centering
\begin{center}
		\includegraphics[width = \columnwidth]{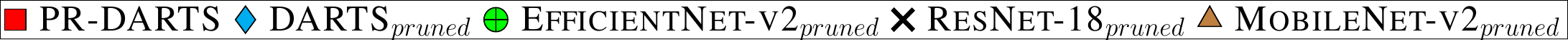}
\end{center}
\resizebox{\columnwidth}{!}{
\begin{tabular}{ccc}
\multicolumn{3}{c}{
}
\\
		\includegraphics[width = \columnwidth]{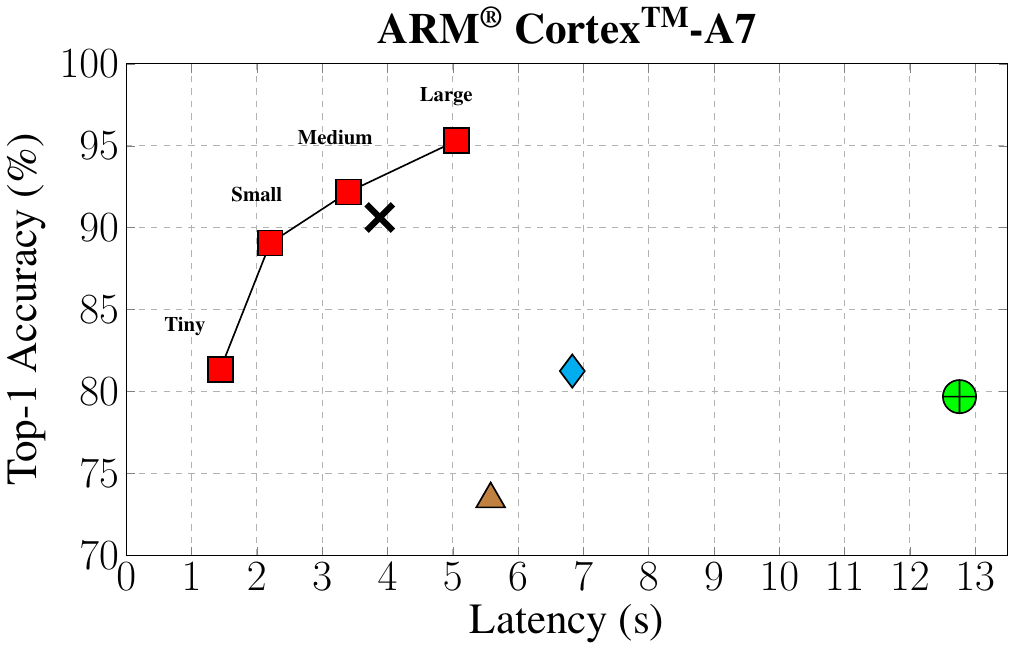}
    &
		\includegraphics[width = \columnwidth]{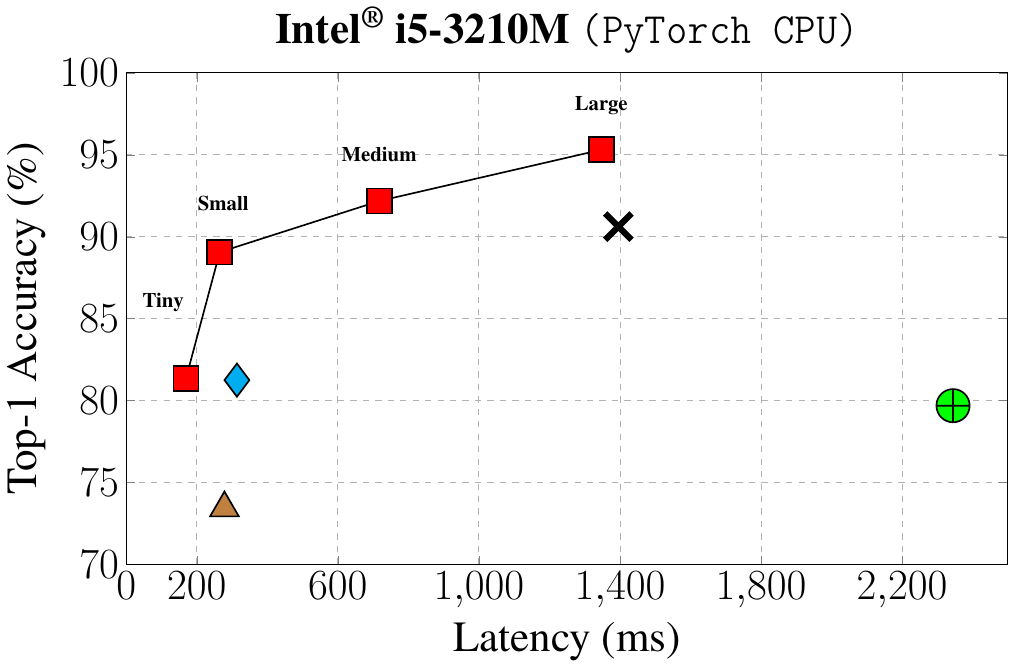}
&
		\includegraphics[width = \columnwidth]{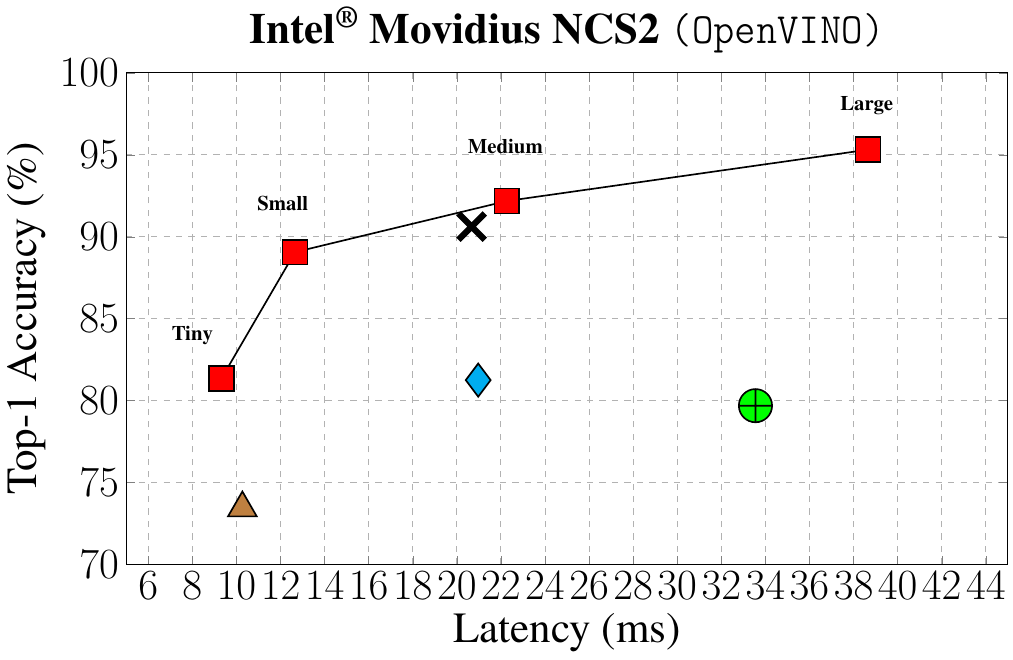}
    \\
		\includegraphics[width = \columnwidth]{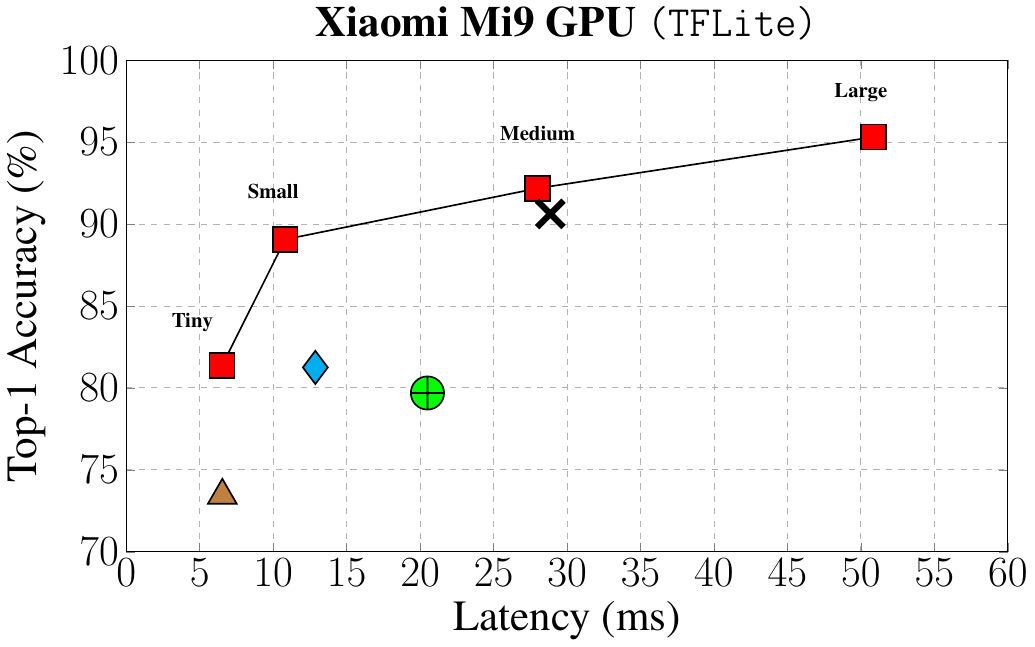}
    &
		\includegraphics[width = \columnwidth]{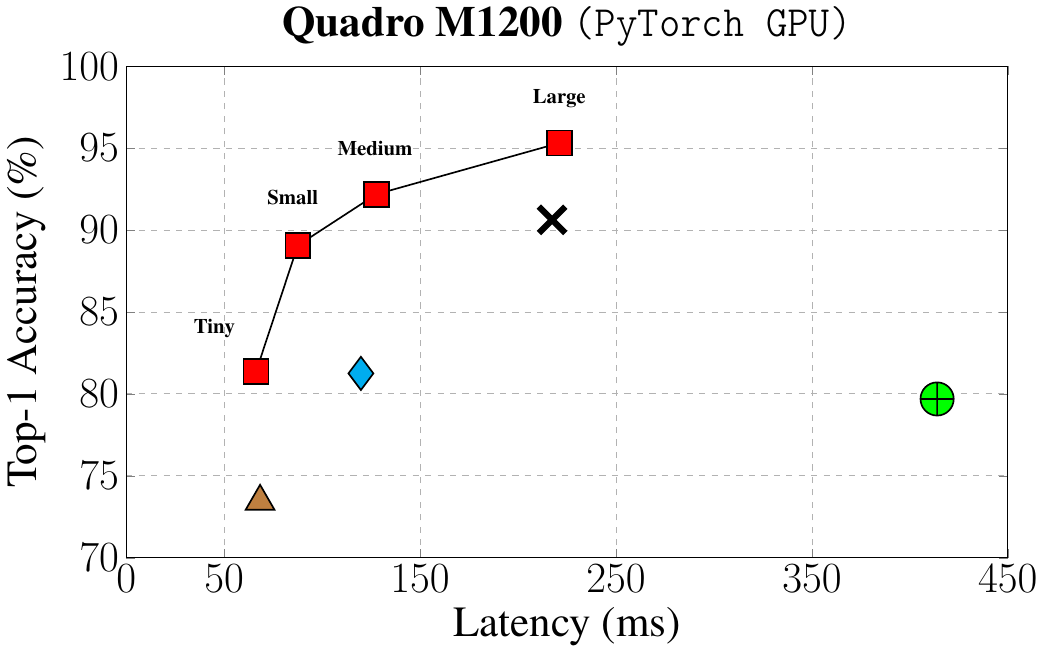}
 &
		\includegraphics[width = \columnwidth]{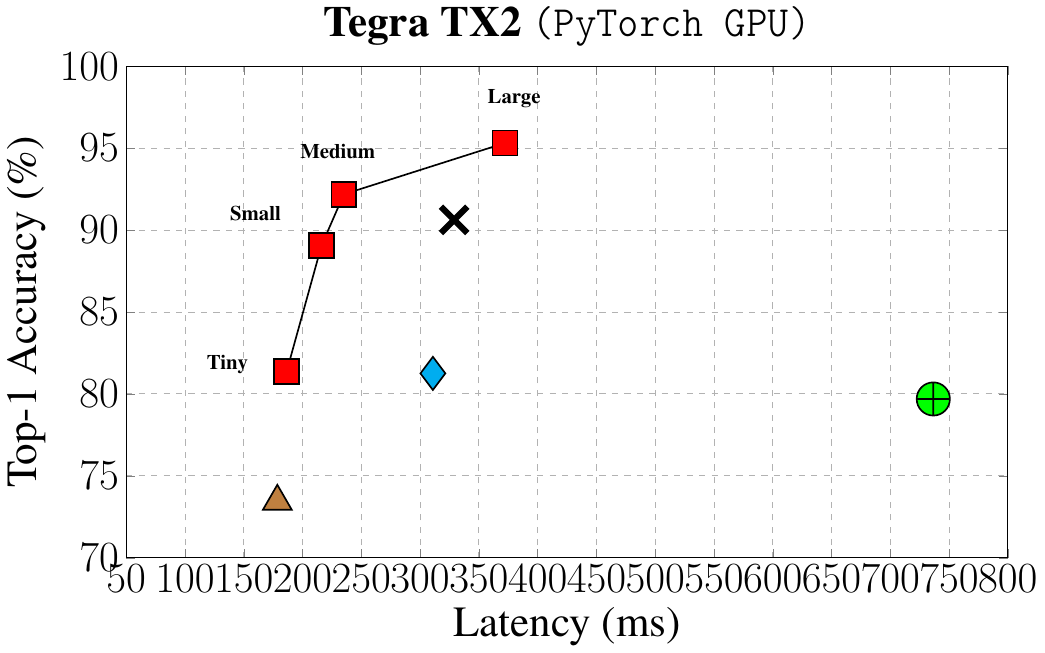}
\end{tabular}}
\captionof{figure}{Trade-off: accuracy v.s. measured latency. DASS-Tiny, DASS-Small, DASS-Medium, DASS-Large are variants of DASS designed for different computational budgets (Table~\ref{tab:ArchitectureConfig}). DASS-Tiny consistently achieves higher accuracy with similar latency than MobileNet-v2$_{sparse}$ and provides lower latency while achieving better accuracy as DARTS$_{sparse}$.}
\label{fig:experiments:hardware_results}
\end{table*}

\subsection{Analyzing the Discrimination Power of DASS}
\label{sec:experiments:analysis}

We use the t-distributed stochastic neighbor embedding (t-SNE) method \citep{van2008visualizing} for visualizing decision boundaries of dense high-performance architecture designed by DARTS, DARTS$_{sparse}$ (sparse dense DART architecture with pruning), and DASS ( our sparse architecture)  on the CIFAR-10 dataset.
Fig.~\ref{fig:experiments:PCA} illustrates the decision boundaries of classification for each network.
According to the results, DASS has a higher discrimination power than DARTS$_{sparse}$, and DASS with a 99\% pruning ratio behaves very similarly to the dense and high-performance DARTS architecture.
\begin{table*}[t]
\captionsetup{justification=centering}
\begin{center}
\begin{center}
		\includegraphics[width = 0.7\columnwidth]{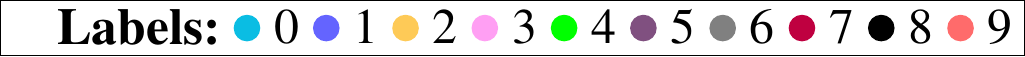}
	\end{center}
\resizebox{\linewidth}{!}{
\begin{tabular}{ccc}
\multicolumn{3}{c}{
}
\\
\includegraphics[width =\columnwidth]{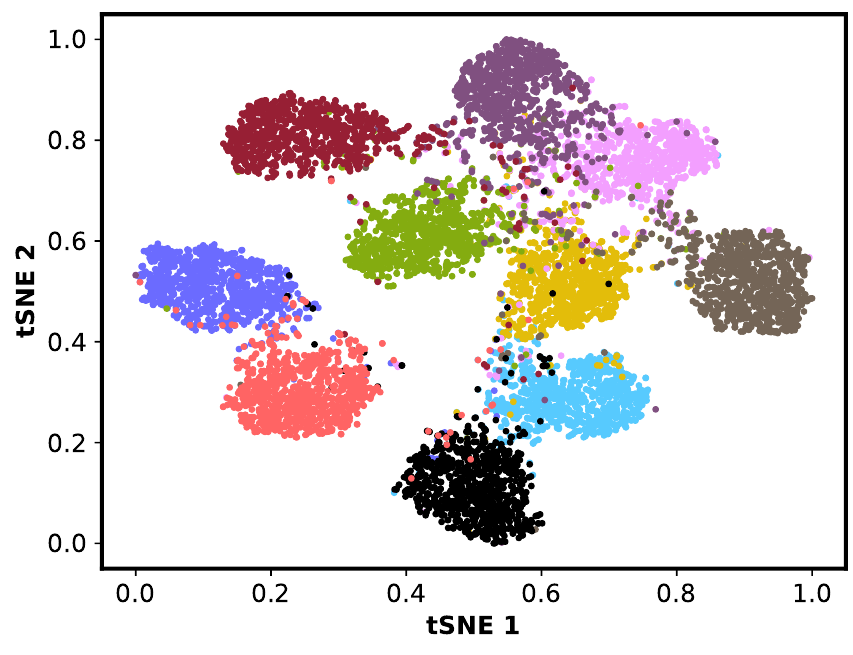}
&
\includegraphics[width =\columnwidth]{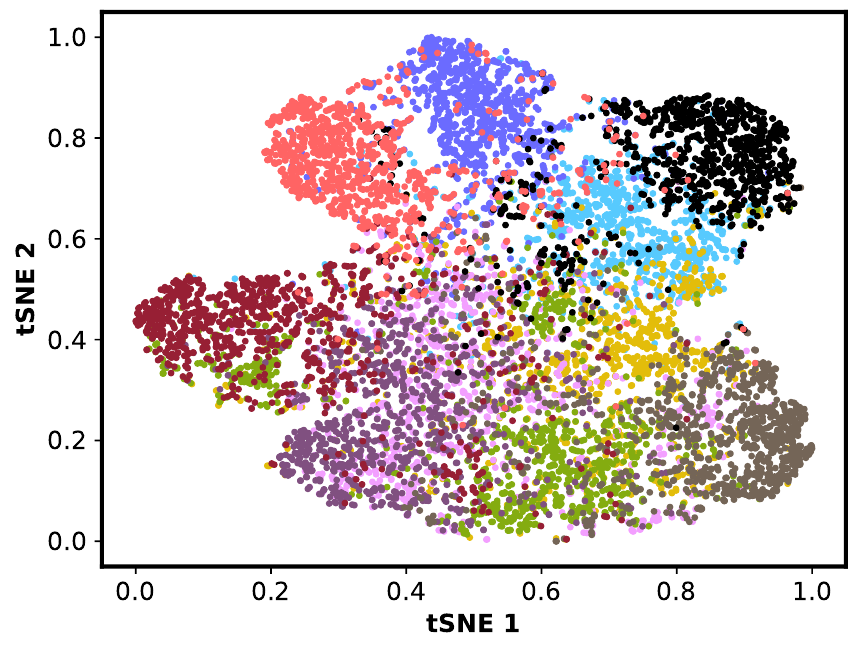}
&
\includegraphics[width =\columnwidth]{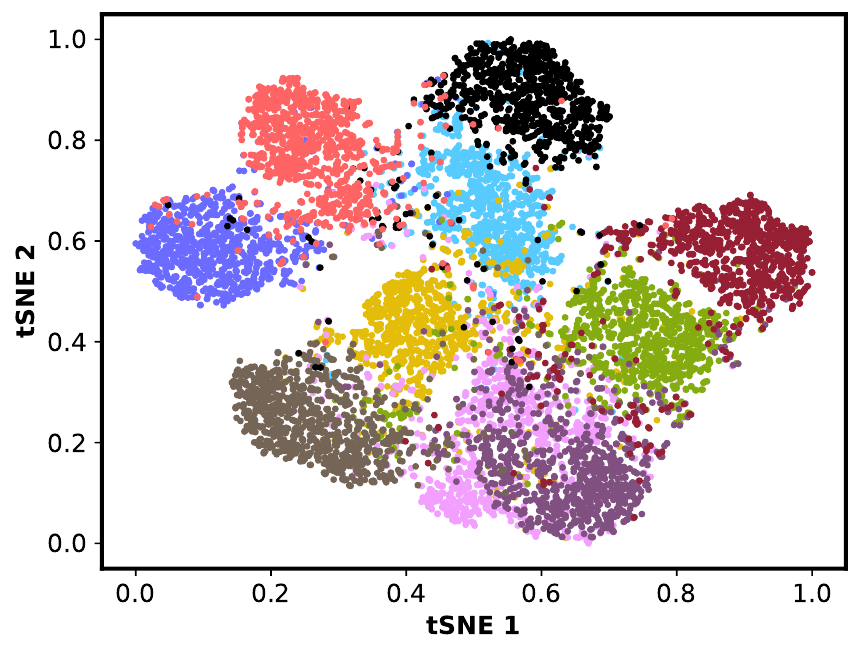}
\\
\textbf{\huge (a) DARTS} &\textbf{ \huge (b) DARTS$_{pruned}$} & \textbf{\huge (c) DASS-Small} 
\end{tabular}}  
\end{center}
\captionof{figure}{Visualize decision boundary of (a) DARTS. (b)  DARTS$_{sparse}$. (c) DASS-Large with t-SNE embedding method.}
\label{fig:experiments:PCA}
\end{table*}

\subsection{Qualitative Analysis of the Searched Cell.}
\label{sec:expriment:qualitativeAnalysis}
\begin{table}[hbpt]
\centering
\resizebox{\columnwidth}{!}{
\begin{tabular}{cc}
\centerline{\includegraphics[width=\columnwidth]{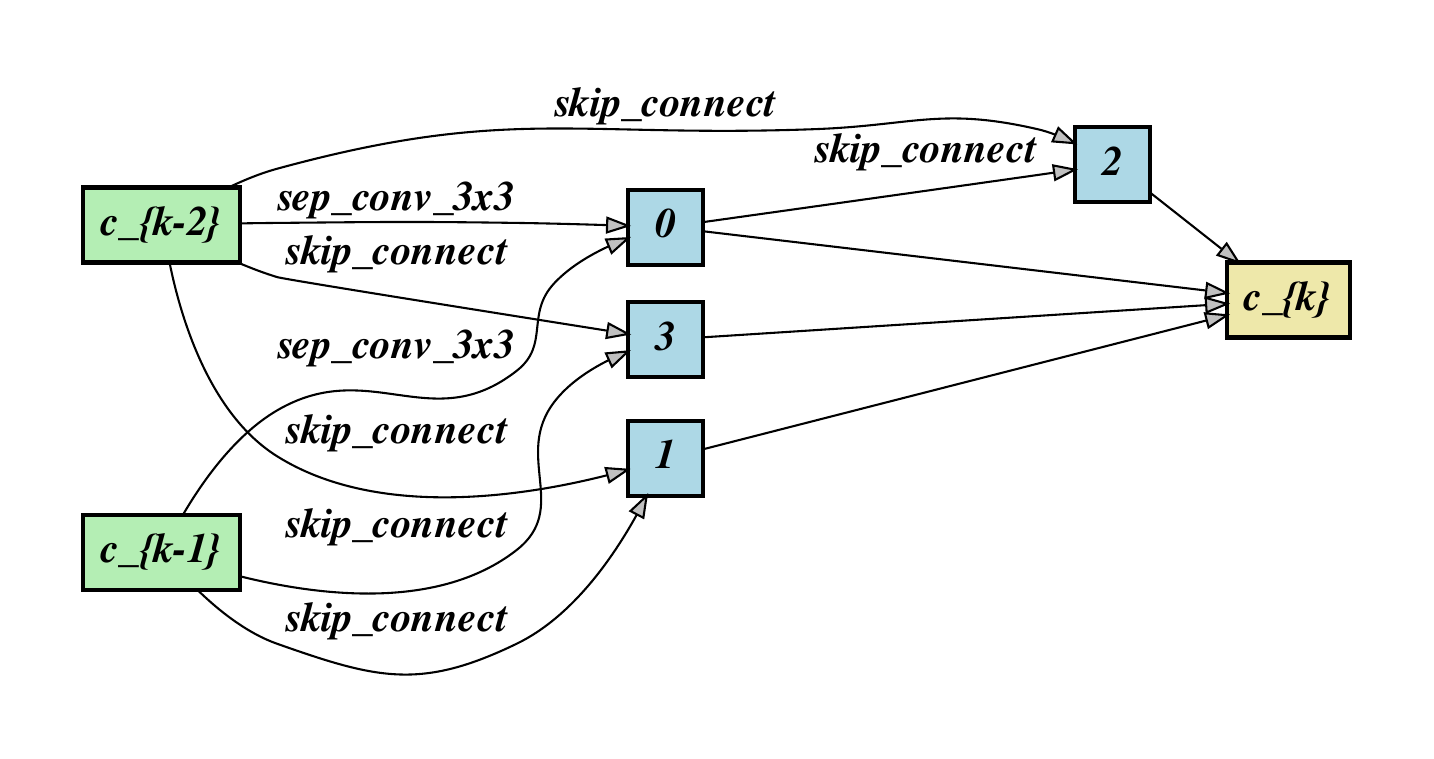}} & \centerline{\includegraphics[width=\columnwidth]{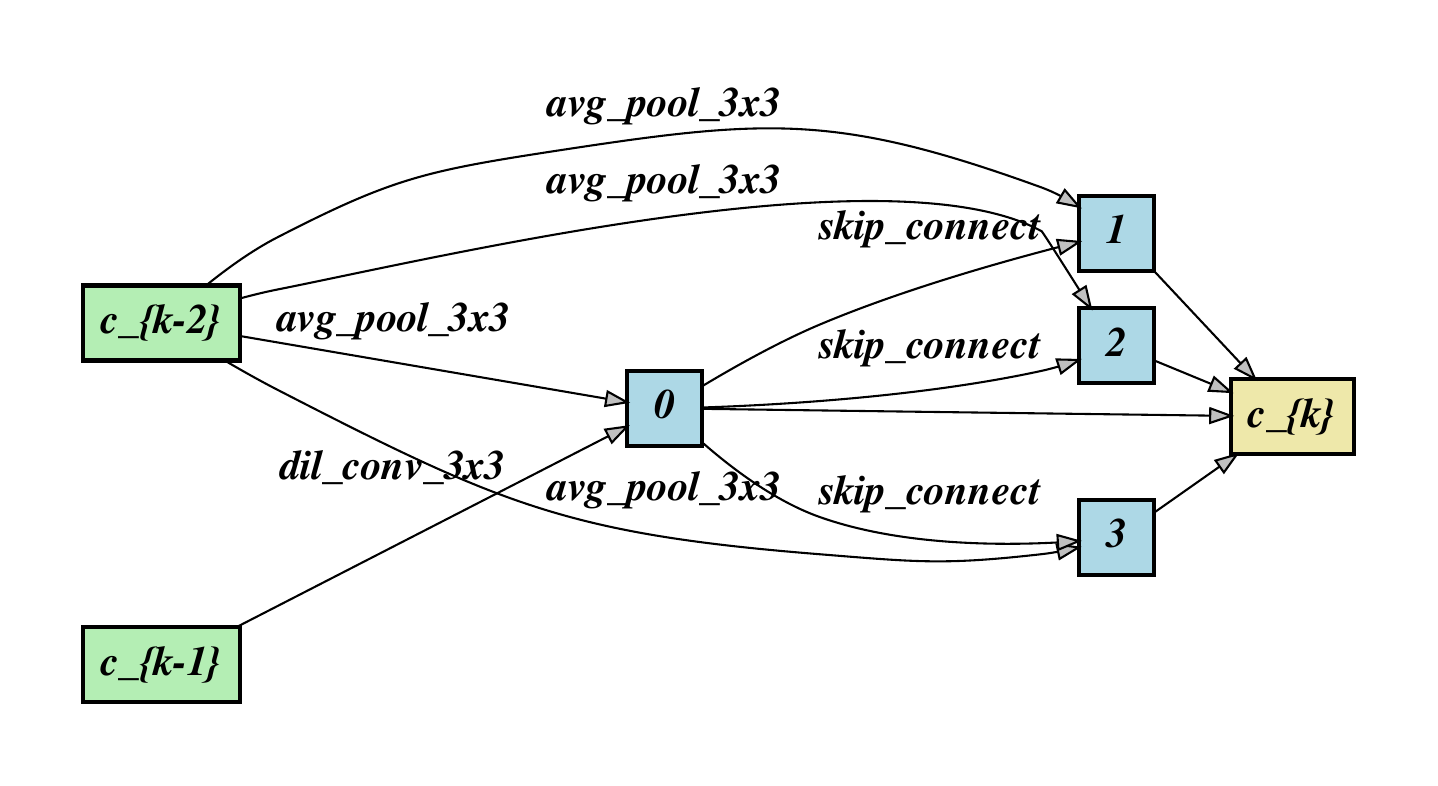}}
\\
\large \textbf{(a) DARTS Normal Cell.} & \large  \textbf{(b) DARTS Reduction Cell.}
\\
\centerline{\includegraphics[width=\columnwidth]{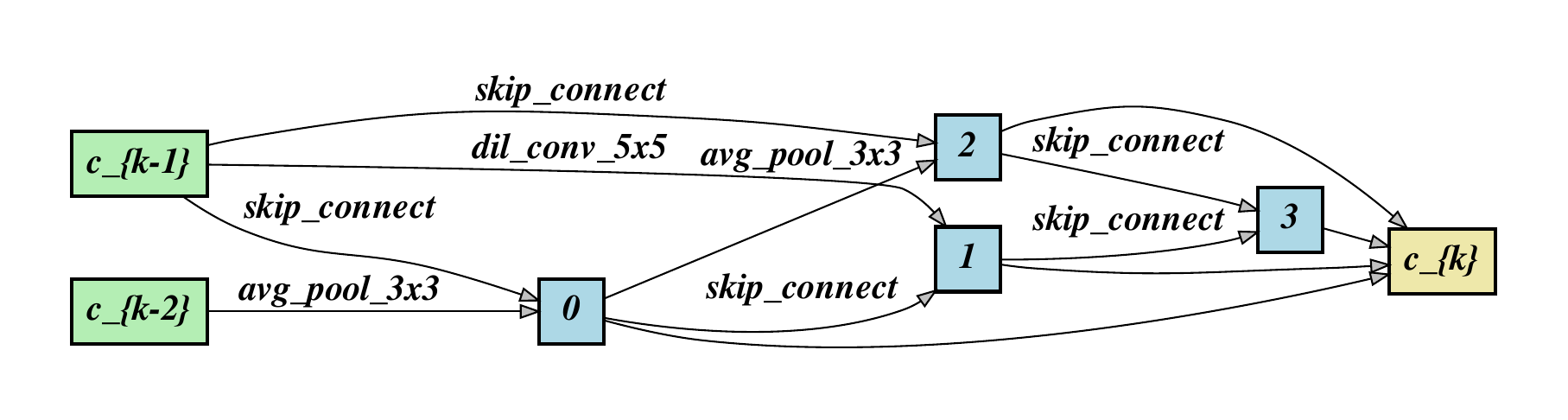}} & \centerline{\includegraphics[width=\columnwidth]{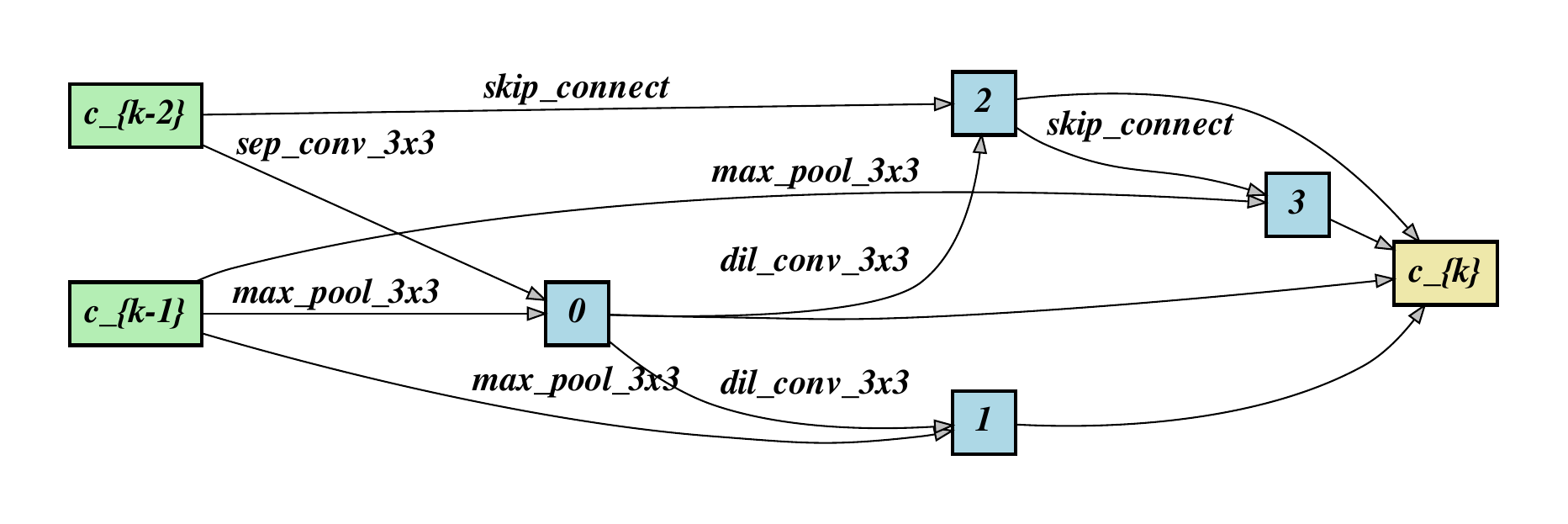}}
\\
\large \textbf{(a) DASS Normal Cell.} & \large  \textbf{(b) DASS Reduction Cell.}
\\

\end{tabular}}
\captionof{figure}{The illustration of (a) normal cell and (b) reduction cell.} 
\label{fig:Supplementary_Material:cell}
\end{table}
Fig.~\ref{fig:Supplementary_Material:cell} shows the best cells searched by DASS-Small.
An interesting finding is that, for the normal cell, DASS-Small tends to select \texttt{SparseConv} operation with larger kernel sizes ($5 \times 5$), providing more pruning candidates to optimize the pruning mask.
DASS-Small tends to leverage max-pooling operations in the reduction cell instead of avg-pooling operations.
This is because the max-pooling operation has a higher feature extraction capability with sparse filters \citep{yu2014mixed}. 
\subsection{Reproducibility Analysis.}
\label{sec:experiments:reproducibility}
\begin{minipage}{0.53\textwidth}

To verify the reproducibility of results, the DASS-Small search procedure was run five times with different random seeds. Fig.~\ref{fig:discussion:reproducibility} plots the average of accuracy and loss variations as well as the shades to indicate the confidence intervals. Results show that, while the confidence interval is wide at first, the average of multiple runs converges to neural architectures with similar performance with an average standard deviation (STDEV) of 2.22\%.
\end{minipage}
\begin{minipage}{0.5\textwidth}
    \begin{center}
		\includegraphics[width = 0.7\columnwidth]{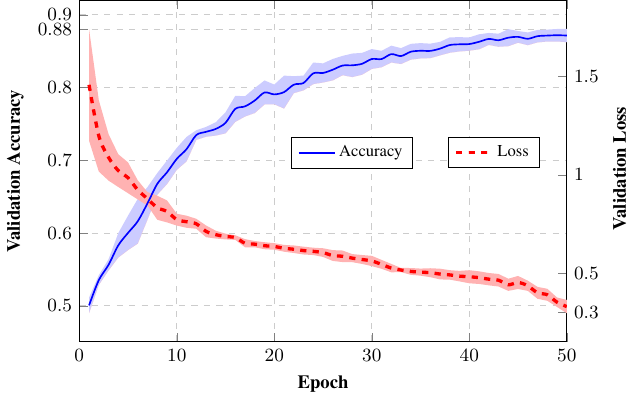}
	\end{center}	
    \label{fig:discussion:reproducibility}
    \captionsetup{hypcap=false}
    \captionof{figure}{Demonstrating the reproducibility of DASS results.}
\end{minipage}
   
\section{Conclusion}
\label{sec:conclusion}

We propose DASS, a differentiable architecture search method, to design high-performance sparse architectures for DNNs.
DASS significantly improves the performance of sparse architectures by proposing: (i) a new search space that contains sparse parametric operations; and (ii) a new search objective that is consistent with sparsity and pruning mechanisms.
Our experimental results reveal that the learned sparse architectures outperform the architectures used in the state-of-the-art on both CIFAR-10 and ImageNet datasets.
In the long term, we foresee that our designed networks can effectively contribute to the goal of green artificial intelligence by efficiently utilizing resource-constrained devices as the edge accelerating solutions. 
A promising avenue for future work is to design a sparse network that is also robust against adversarial attacks.

\bibliographystyle{unsrt}

\end{document}